\documentclass[sigconf, screen]{acmart}
\usepackage[nowarn]{glossaries}
\makeglossaries
\newacronym{test}{TEST}{test}
\newacronym{sd}{SD}{Standard deviation}

\newacronym{semg}{sEMG}{Surface Electromyography}
\newacronym{emg}{EMG}{Electromyography}
\newacronym{fmg}{FMG}{Force Myography}
\newacronym{fsr}{FSR}{Force Sensitive Resistor}

\newacronym{fcnn}{FCNN}{Fully-Connected Neural Network}
\newacronym{lda}{LDA}{Linear Discriminant Analysis}
\newacronym{svm}{SVM}{Support Vector Machine}
\newacronym{cnn}{CNN}{Convolutional Neural Network}
\newacronym{dnn}{DNN}{Deep Neural Network}

\newacronym{relu}{ReLU}{Rectified Linear Unit}

\newacronym{pnn}{PNN}{Progressive Neural Networks}
\newacronym{cpnn}{CPNN}{Combined Progressive Neural Networks}

\newacronym{hp}{HP}{Hyperparameter}
\newacronym{hpo}{HPO}{Hyperparameter Optimisation}
\glsdisablehyper
\usepackage[nameinlink,noabbrev]{cleveref}
\usepackage{graphicx}
\usepackage{caption}
\usepackage{subcaption}
\usepackage{siunitx}
\renewcommand\footnotetextcopyrightpermission[1]{} 

\AtBeginDocument{%
  \providecommand\BibTeX{{%
    \normalfont B\kern-0.5em{\scshape i\kern-0.25em b}\kern-0.8em\TeX}}}

\begin{document}

\title{Force myography benchmark data for hand gesture recognition and transfer learning}

\author{Thomas Buhl Andersen, R\'ogvi Eliasen, Mikkel Jarlund, Bin Yang}
\affiliation{%
    \institution{Department of Computer Science, Aalborg University, Denmark}
    tbuhla@gmail.com \hspace{5pt}
    rogvieliasen@gmail.com \hspace{5pt} mikkel.jarlund@gmail.com \hspace{5pt}
    byang@cs.aau.dk
}






\newcommand{\varnumsubjects}{20}
\newcommand{\varmale}{17}
\newcommand{\varfemale}{3}

\newcommand{\varmeanage}{23.85}
\newcommand{\varsdage}{1.53}
\newcommand{\varmeanexercise}{2.35}
\newcommand{\varsdexercise}{1.79}
\newcommand{\varinjured}{1}
\newcommand{\varnoninjured}{19}
\newcommand{\varmeanwrist}{17.86}
\newcommand{\varsdwrist}{1.07}
\newcommand{\varmeanforearm}{26.9}
\newcommand{\varsdforearm}{1.66}
\newcommand{\varrighthanded}{18}
\newcommand{\varlefthanded}{2}
\newcommand{\varnumgestures}{18}
\newcommand{\vargestureduration}{5 seconds}
\newcommand{\varnumrepetitions}{5}
\newcommand{\varfrequency}{\SI{975}{\hertz}-\SI{1000}{\hertz}}
\newcommand{\varwristposition}{5 cm}

\newcommand{\varforearmposition}{5 cm}
\newcommand{\varwristbandlength}{22 cm}
\newcommand{\varforearmbandlength}{27 cm}

\begin{abstract}
Force myography 
has recently gained increasing attention for hand gesture recognition tasks. However, there is a lack of publicly available benchmark data, with most existing studies collecting their own data often with custom hardware and for varying sets of gestures. 
This limits the ability to compare various algorithms, as well as the possibility for research to be done without first needing to collect data oneself. We contribute to the advancement of this field by making accessible a benchmark dataset collected using a commercially available sensor setup from \varnumsubjects{} persons covering \varnumgestures{} unique gestures, in the hope of allowing further comparison of results as well as easier entry into this field of research. 
We illustrate one use-case for such data, showing how we can improve gesture recognition accuracy by utilising transfer learning to incorporate data from multiple other persons. This also illustrates that the dataset can serve as a benchmark dataset to facilitate research on transfer learning algorithms. 
\end{abstract}



\settopmatter{printfolios=true}

\maketitle
 \pagestyle{plain} 

\section{Introduction}
Gesture recognition is increasingly becoming more popular, as it can be used in various fields, such as rehabilitation in healthcare, smart-homes where gestures can be used as commands, and prosthetic limbs~\cite{health_care, smart_home, bionic_arm}. Different technologies can be used for collecting gesture recognition data, most common among them is \gls{semg} which measures the electrical signals when activating the muscles and \gls{fmg} which measures the mechanical activity of the muscles, i.e., how a muscle changes shape when it is used~\cite{FMG_survey}.

Recent studies using \gls{fmg} often collect their own data, e.g., different sets of gestures, using their own custom hardware~\cite{fmg_data_collection_1, fmg_data_collection_2, fmg_data_collection_3, fmg_data_collection_4}. This creates a reproducibility problem where the custom hardware settings cannot be exactly configured by other researchers and the results obtained on different setups cannot be effectively compared to each other. 

We argue that using a commercially available product eases the hardware configuration issues and publishing a \gls{fmg} benchmark dataset collected from the product makes the comparison of different results easier. In addition, having a publicly available dataset allows for easier entry into the field of hand gesture recognition using \gls{fmg} as it will not require everyone entering the field to first collect their own data. As such we have used a commercially available product BIOX Armbands\footnote{\url{https://www.bioxgroup.dk/product/biox-armband/}} for collecting data from \varnumsubjects{} persons, each person covering \varnumgestures{} unique gestures. The collected data is made publicly available~\cite{data} such that those entering the field of \gls{fmg} based hand gesture recognition can use this data during their research, and also contribute to the dataset. 

In addition, in this paper we show a use case of the dataset on hand gesture recognition using transfer learning~\cite{transfer_learning_survey} and multi-task learning~\cite{DBLP:conf/cikm/Kieu0GJ18}. Here, different persons are considered as different domains (or tasks) and we hope to use data from other persons, i.e., other domains (or tasks), to improve the accuracy of hand gesture recognition of a person. More specifically, we train a baseline \gls{fcnn} only using the data from a specific person, and a \gls{cpnn} model also using data from different persons.  We then evaluate the improvement of using transfer learning. The source code is also available in the same GitHub repository. By doing this, we hope to illustrate that the dataset not only contributes to hand gesture recognition, an increasingly important application area, but can also facilitate research on transfer learning, a significant research direction in machine learning. 

\section{Related work}\label{sec:related}
 We review four independent research papers that utilise \gls{fmg} data to uncover their data collection process. Henceforth we will refer to the people that data has been collected for as subjects.

In the first paper we examined~\cite{fmg_data_collection_1}, data was collected using a custom armband with \SI{8}{sensors} wrapped around the upper forearm. Data was collected for \SI{10}{subjects} each performing \SI{6}{gestures} with a sampling frequency of \SI{10}{\hertz}. In the second paper, data was collected using a custom wristband with \SI{15}{sensors} wrapped around the wrist fraction of the arm. Data was collected for \SI{10}{subjects} each performing \SI{6}{gestures} \SI{12}{times} with a sampling frequency of \SI{30}{\hertz}~\cite{fmg_data_collection_2}. In the third paper, data was collected using a custom armband with \SI{16}{sensors} wrapped around the forearm fraction of the arm. Data was collected for \SI{12}{subjects} each performing \SI{48}{gestures} \SI{5}{times} with a sampling frequency of \SI{10}{\hertz}~\cite{fmg_data_collection_4}. In the fourth paper, data was collected using a custom armband with \SI{16}{sensors} on the dorsal side and \SI{16}{sensors} on the volar side of the forearm. Data was collected for \SI{6}{subjects} each performing \SI{17}{gestures} \SI{4}{times} with a sampling frequency of \SI{100}{\hertz}~\cite{fmg_data_collection_3}. 

In summary of the related work, we observe that \gls{fmg} data collection varies widely in method and execution, making it hard to perform any kind of meaningful method comparisons to assess the performances of different methods and to identify the state-of-the-art methods. Our data set includes sensors for both the wrist and forearm, specifically, 7 sensors for the wrist and 8 sensors for the forearm. The data set is collected  at up to 1000~Hz from \varnumsubjects{} subjects, covering \varnumgestures{} unique gestures. Thus, it is possible to sample our data set to derive subsets, e.g., which represent data sets that are collected with a lower frequency and data sets that are with less gestures and subjects. 
\section{Data collection}\label{sec:data_collection}
We have collected data from a total of \varnumsubjects{} subjects. We will here describe the collected data for each subject as well as the protocol we followed during our data collection process. For each subject we collected contextual information, fitted the sensors, performed calibration and collected sensor readings.

\begin{figure*}[hbt!]
    \centering
    \captionsetup{justification=centering}
    \begin{minipage}[t]{0.16\linewidth}
        \includegraphics[width=\textwidth]{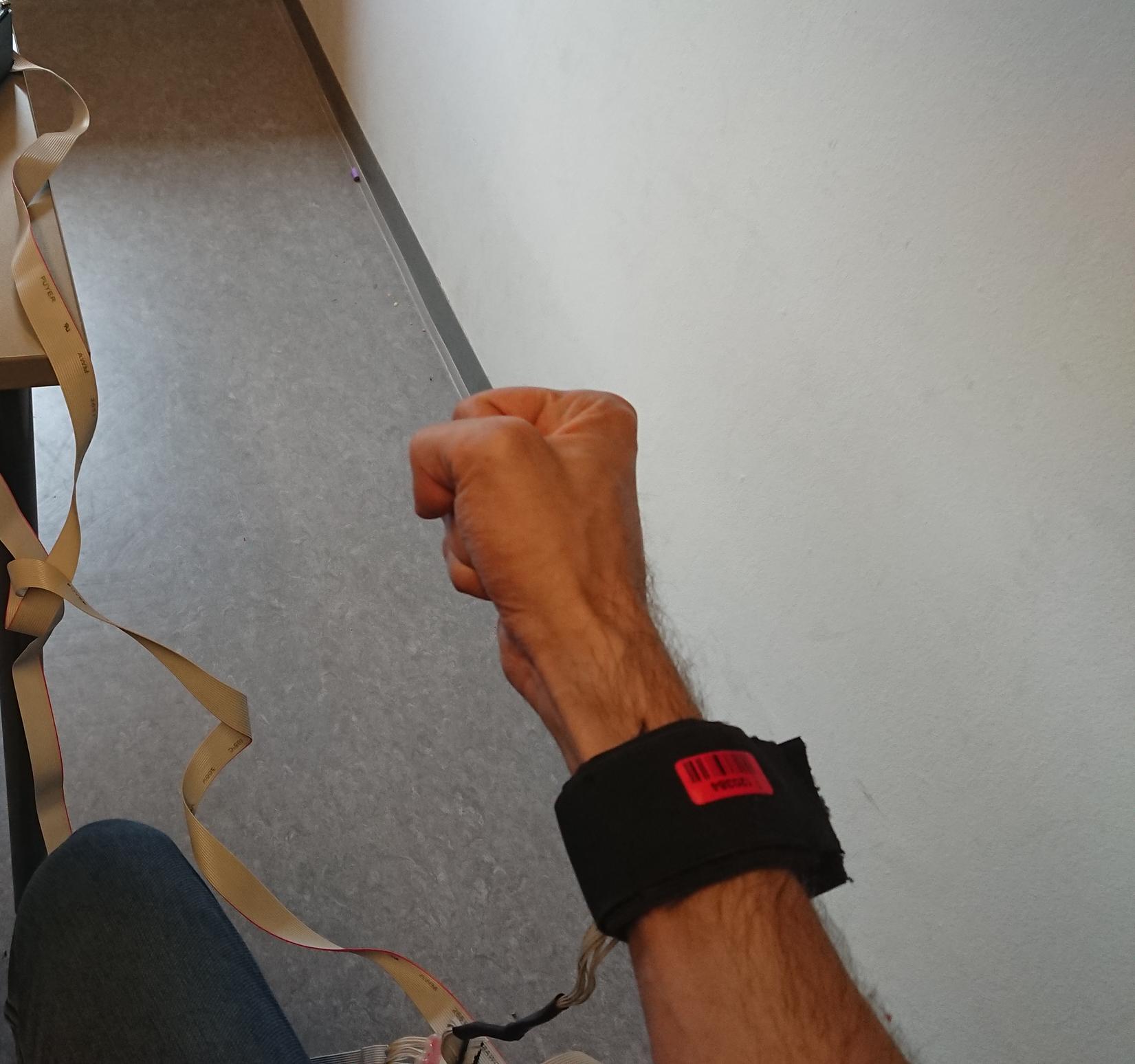}
        \subcaption{\\Neutral closed}
        \label{fig:neutral_closed}
    \end{minipage}
    \begin{minipage}[t]{0.16\linewidth}
        \includegraphics[width=\textwidth]{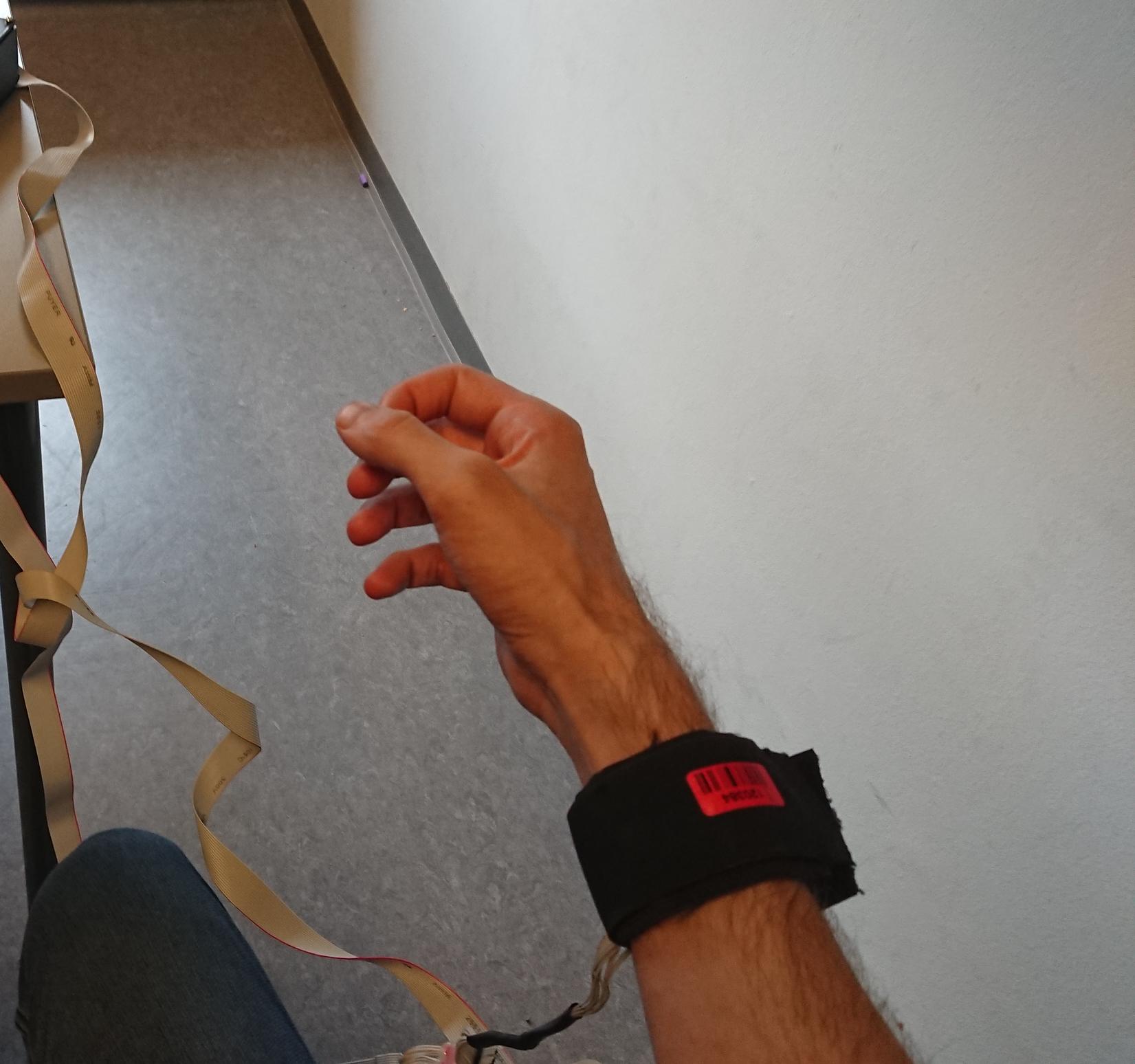}
        \subcaption{\\Neutral rest}
        \label{fig:neutral_rest}
    \end{minipage}
    \begin{minipage}[t]{0.16\linewidth}
        \includegraphics[width=\textwidth]{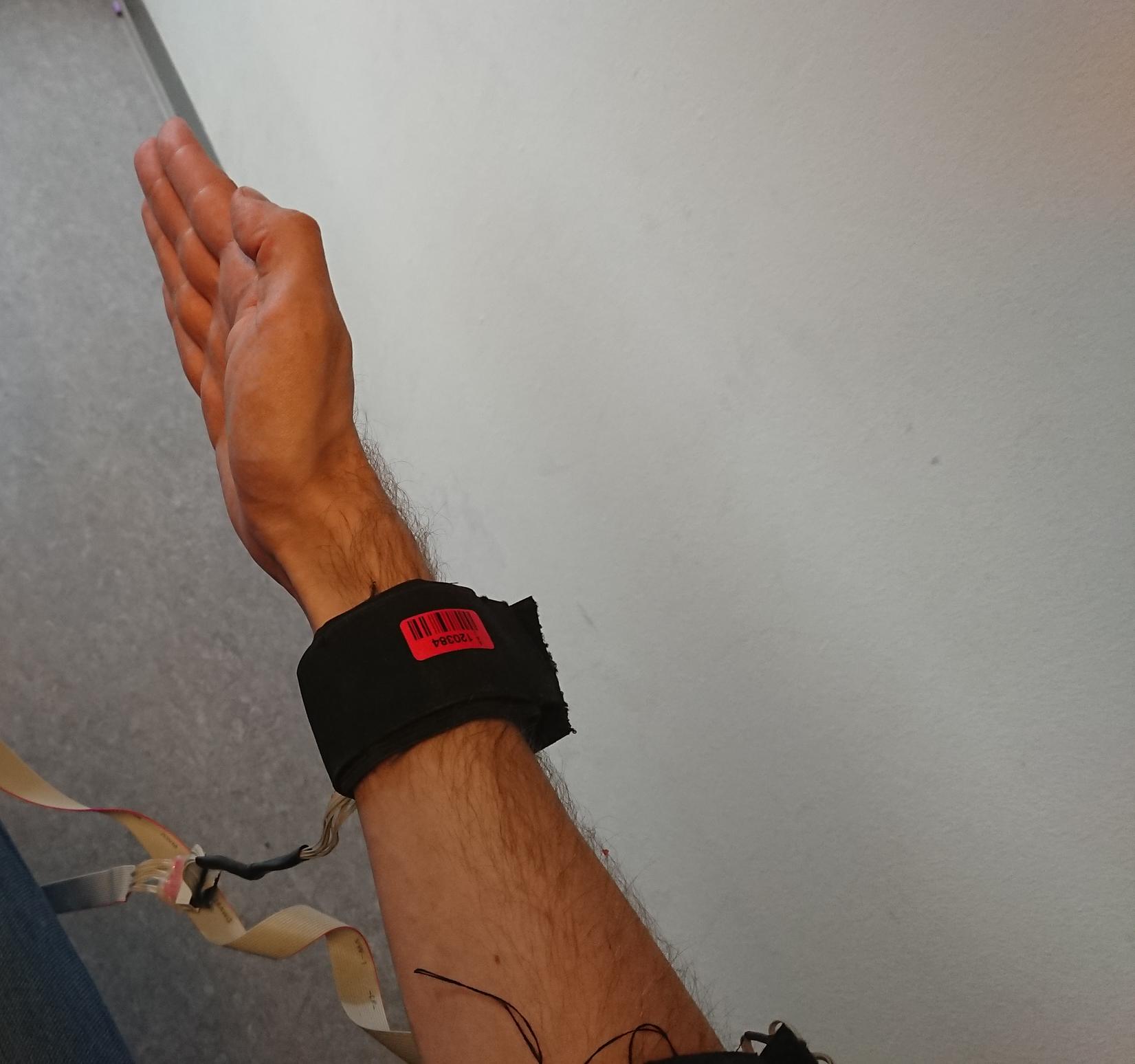}
        \subcaption{\\Neutral open}
        \label{fig:neutral_straight}
    \end{minipage}
    \begin{minipage}[t]{0.16\linewidth}
        \includegraphics[width=\textwidth]{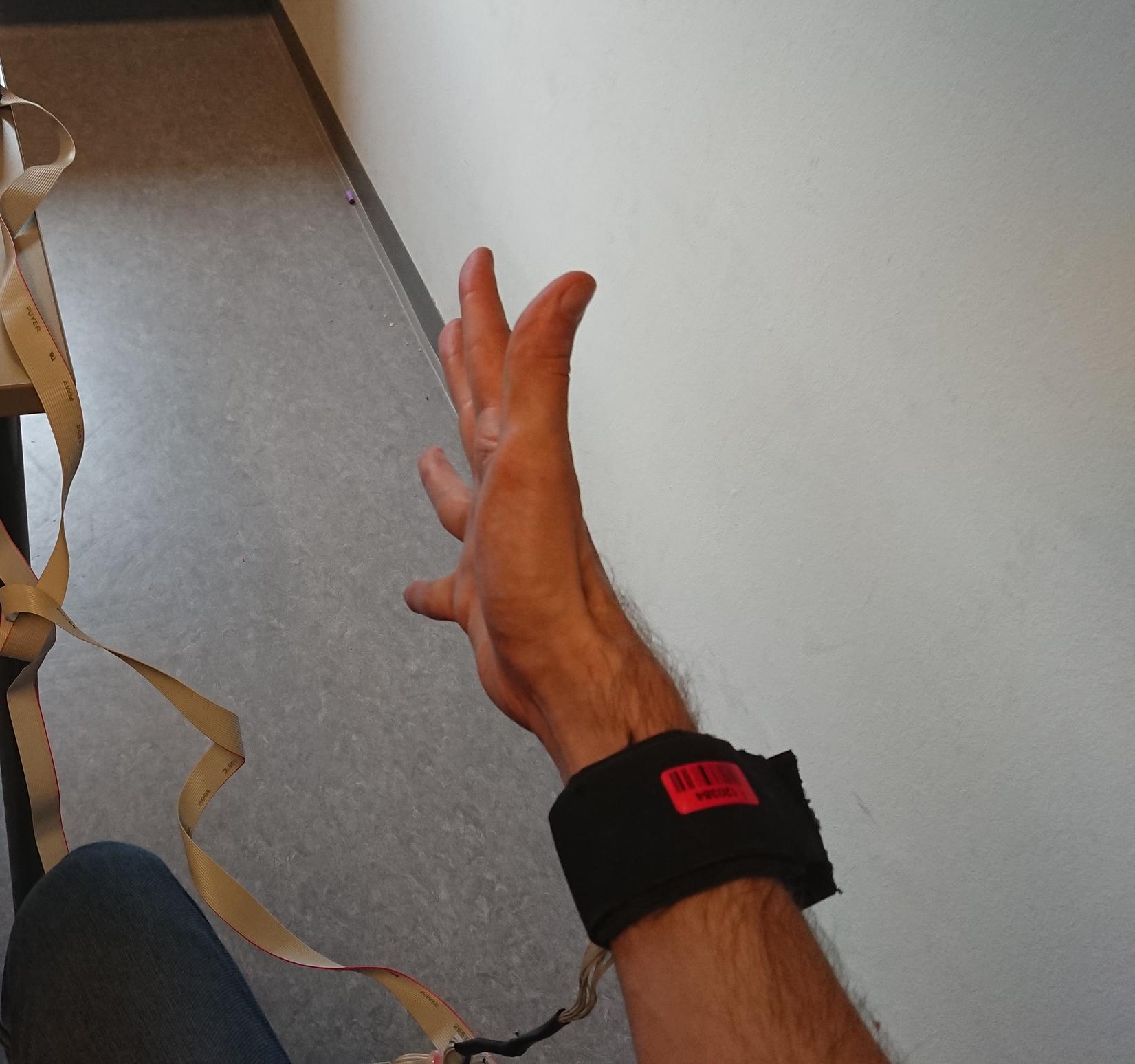}
        \subcaption{\\Neutral wide}
        \label{fig:neutral_wide}
    \end{minipage}
    \begin{minipage}[t]{0.16\linewidth}
        \includegraphics[width=\textwidth]{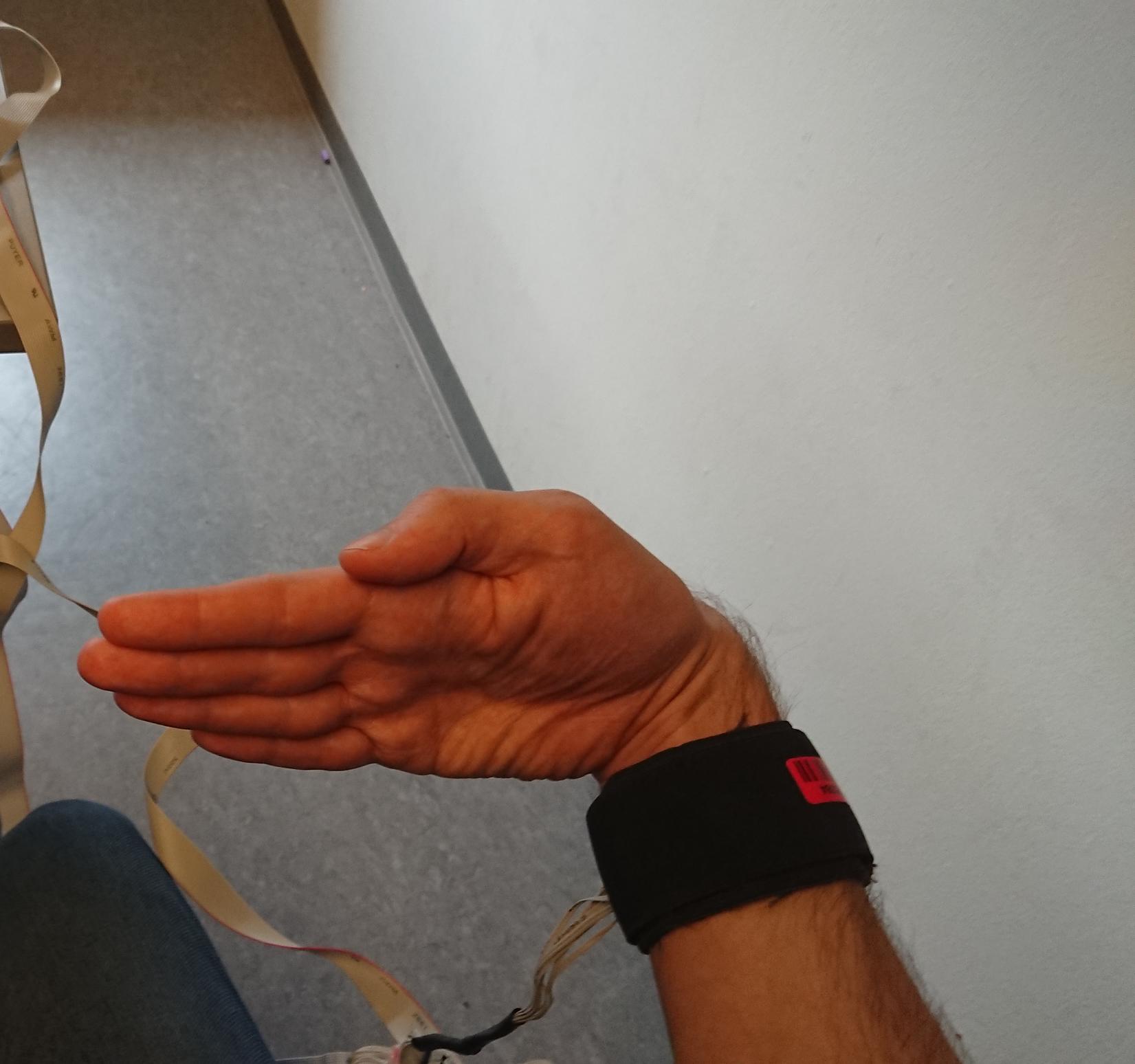}
        \subcaption{\\Neutral flexion}
        \label{fig:neutral_flexion}
    \end{minipage}
    \begin{minipage}[t]{0.16\linewidth}
        \includegraphics[width=\textwidth]{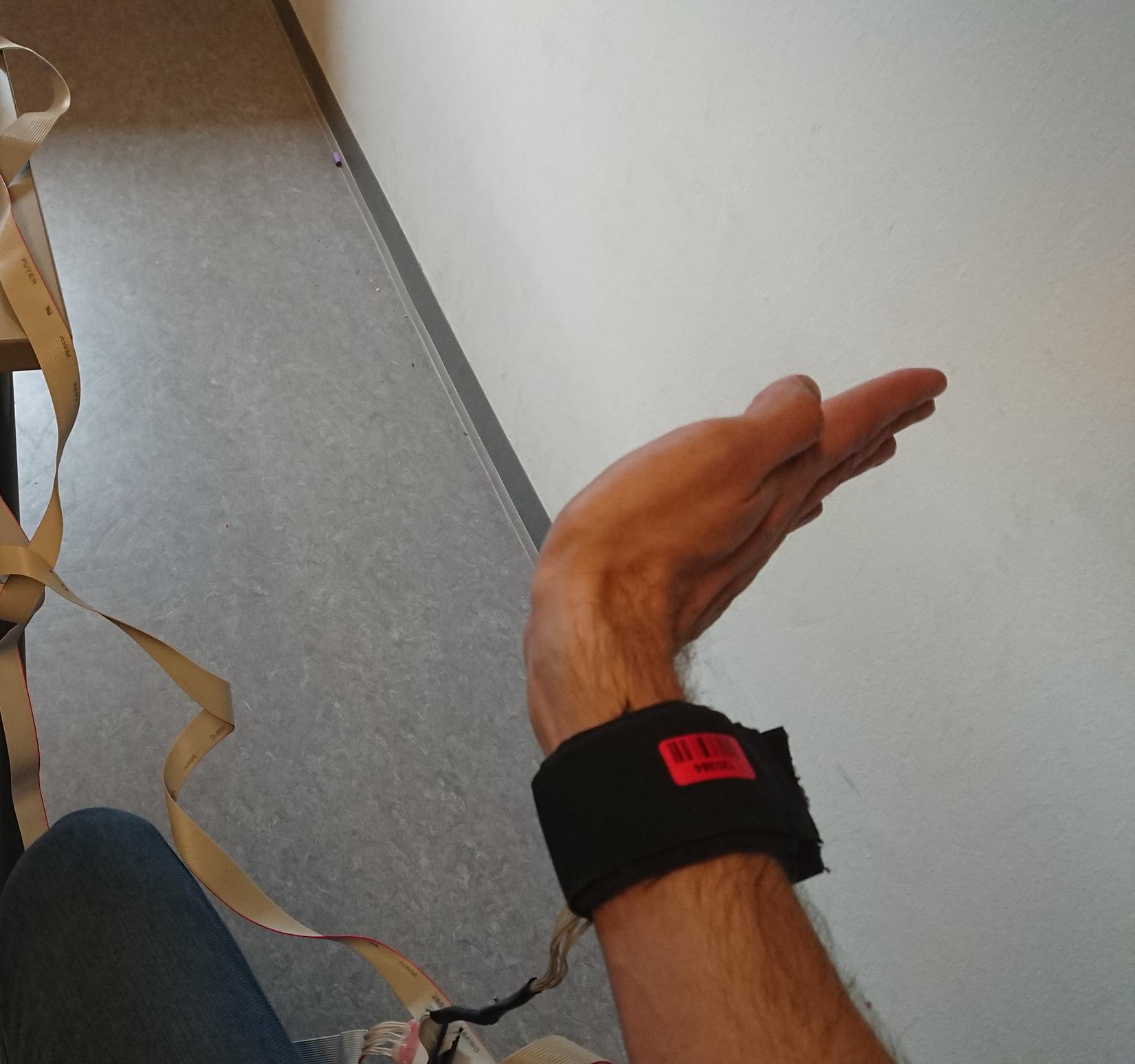}
        \subcaption{\\Neutral extension}
        \label{fig:neutral_extension}
    \end{minipage}
    \\[0.25cm]
    \begin{minipage}[t]{0.16\linewidth}
        \includegraphics[width=\textwidth]{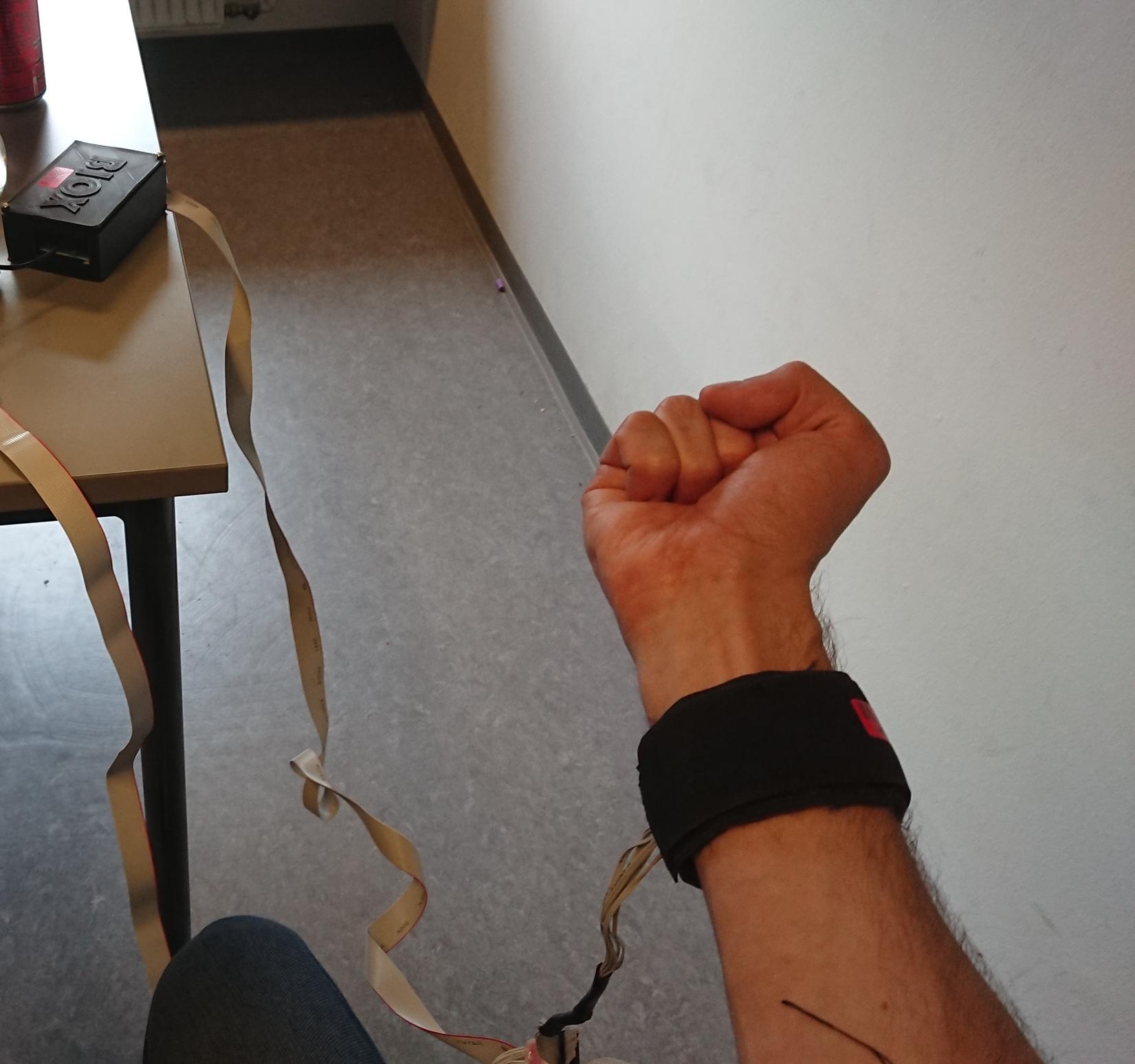}
        \subcaption{\\Supine closed}
        \label{fig:supine_closed}
    \end{minipage}
    \begin{minipage}[t]{0.16\linewidth}
        \includegraphics[width=\textwidth]{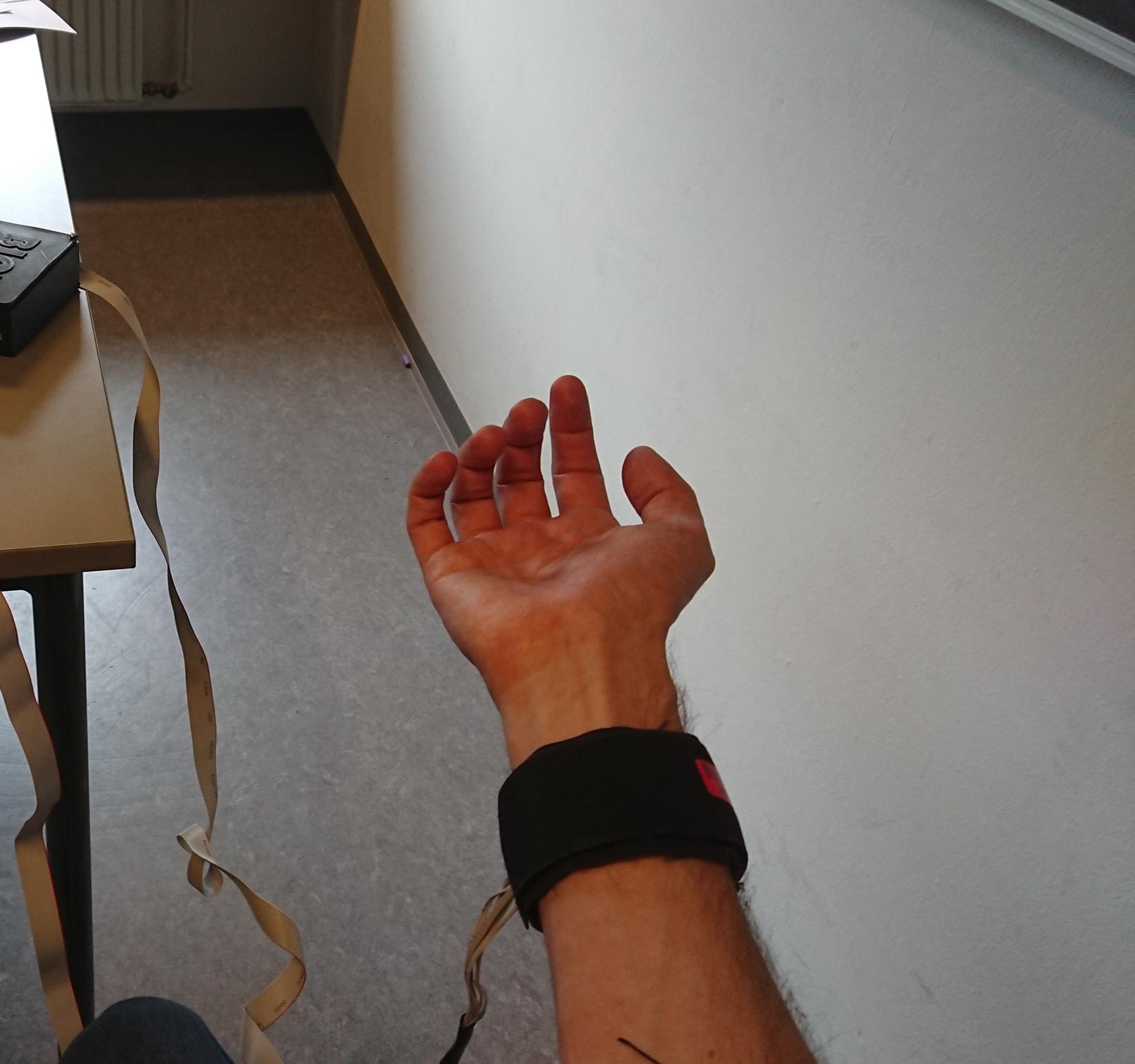}
        \subcaption{\\Supine rest}
        \label{fig:supine_rest}
    \end{minipage}
    \begin{minipage}[t]{0.16\linewidth}
        \includegraphics[width=\textwidth]{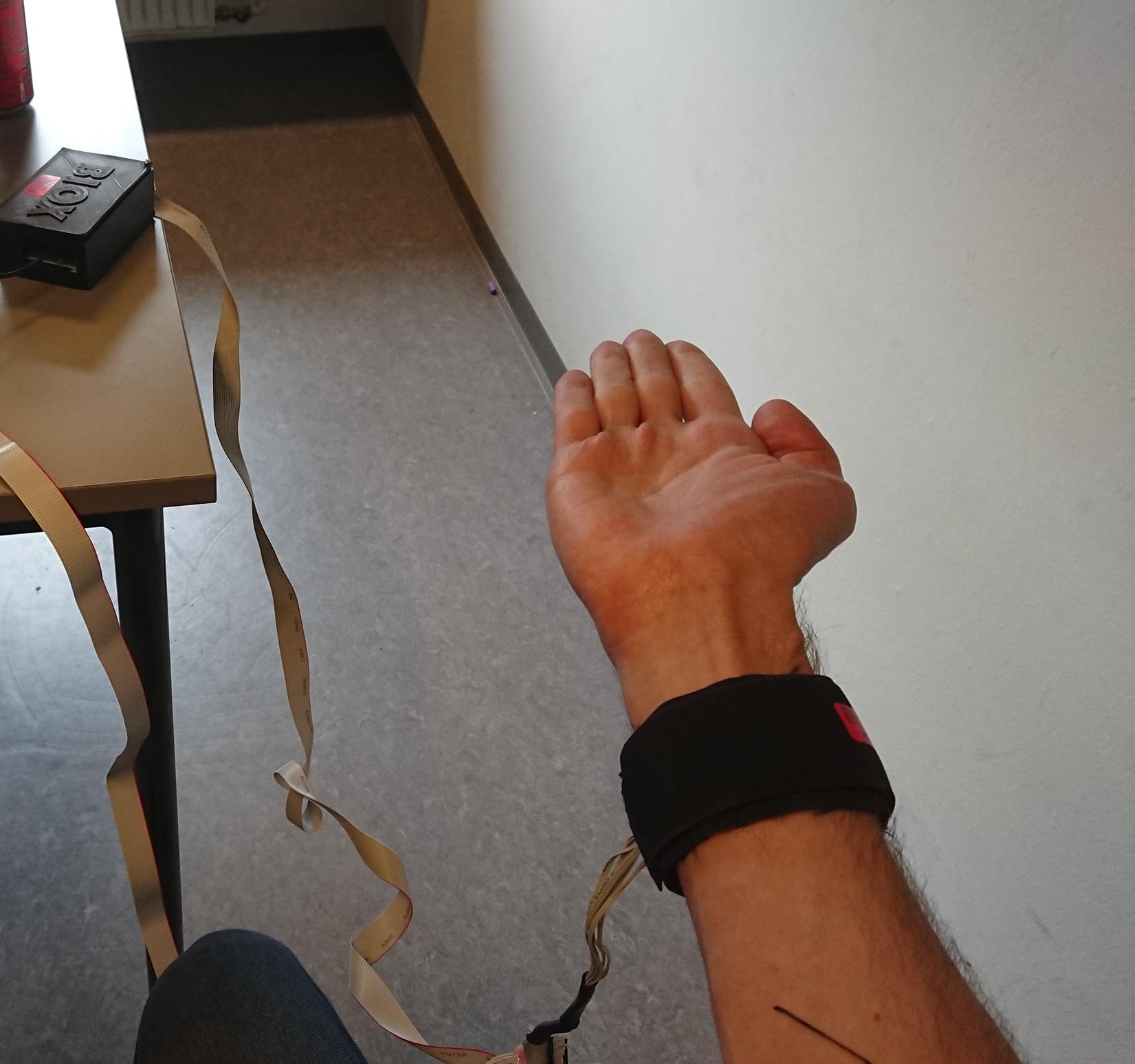}
        \subcaption{\\Supine open}
        \label{fig:supine_straight}
    \end{minipage}
    \begin{minipage}[t]{0.16\linewidth}
        \includegraphics[width=\textwidth]{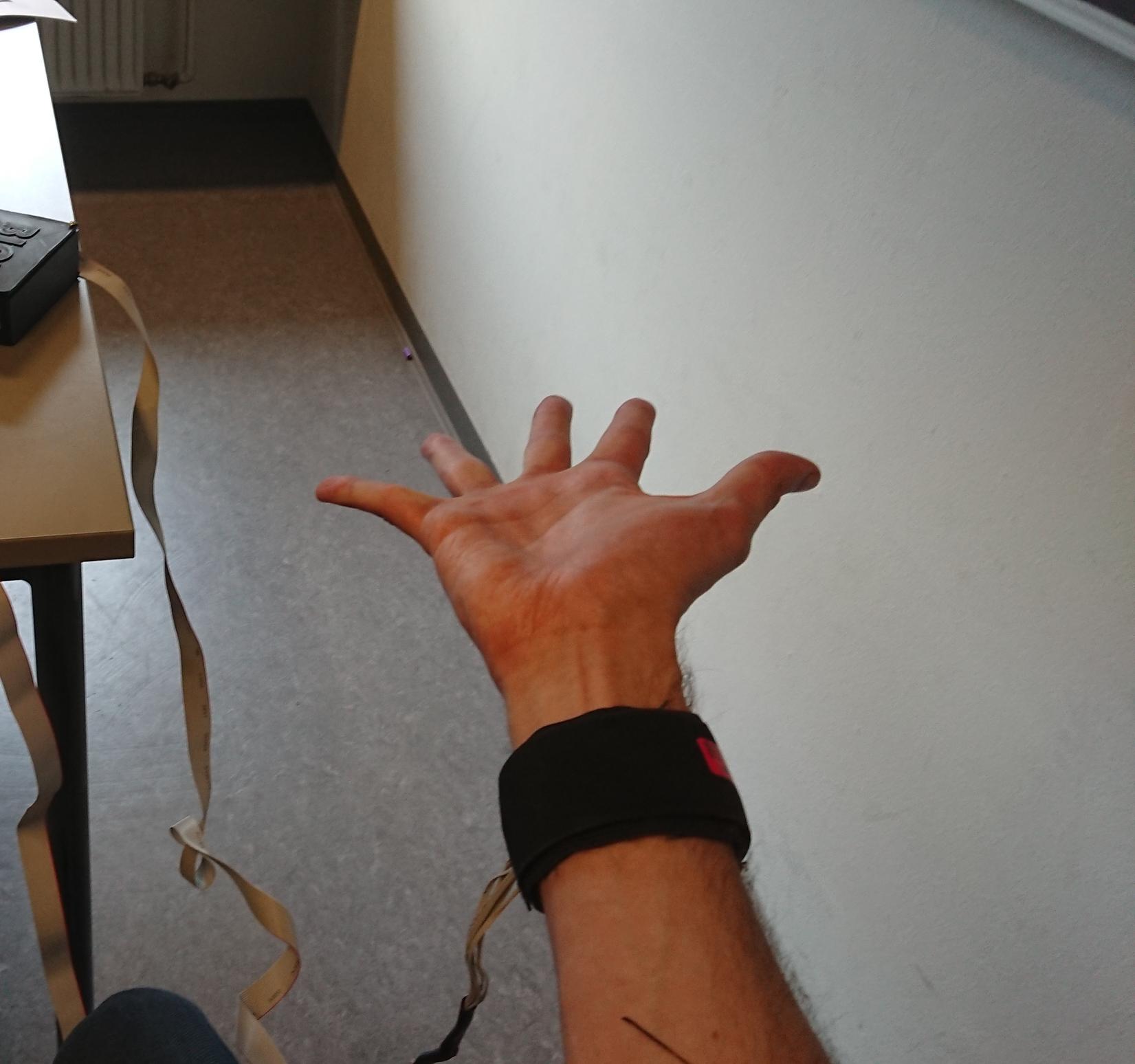}
        \subcaption{\\Supine wide}
        \label{fig:supine_wide}
    \end{minipage}
    \begin{minipage}[t]{0.16\linewidth}
        \includegraphics[width=\textwidth]{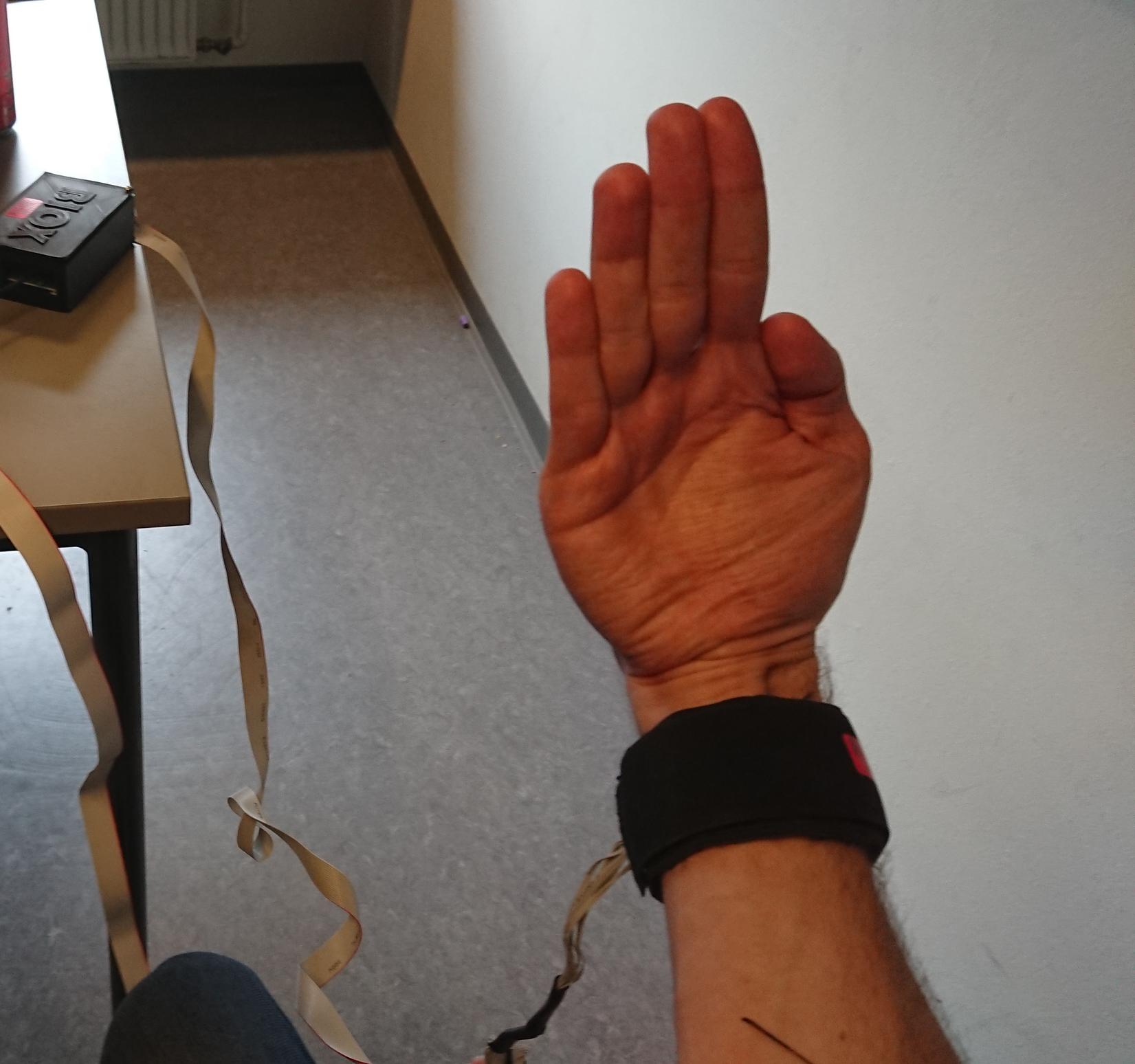}
        \subcaption{\\Supine flexion}
        \label{fig:supine_flexion}
    \end{minipage}
    \begin{minipage}[t]{0.16\linewidth}
        \includegraphics[width=\textwidth]{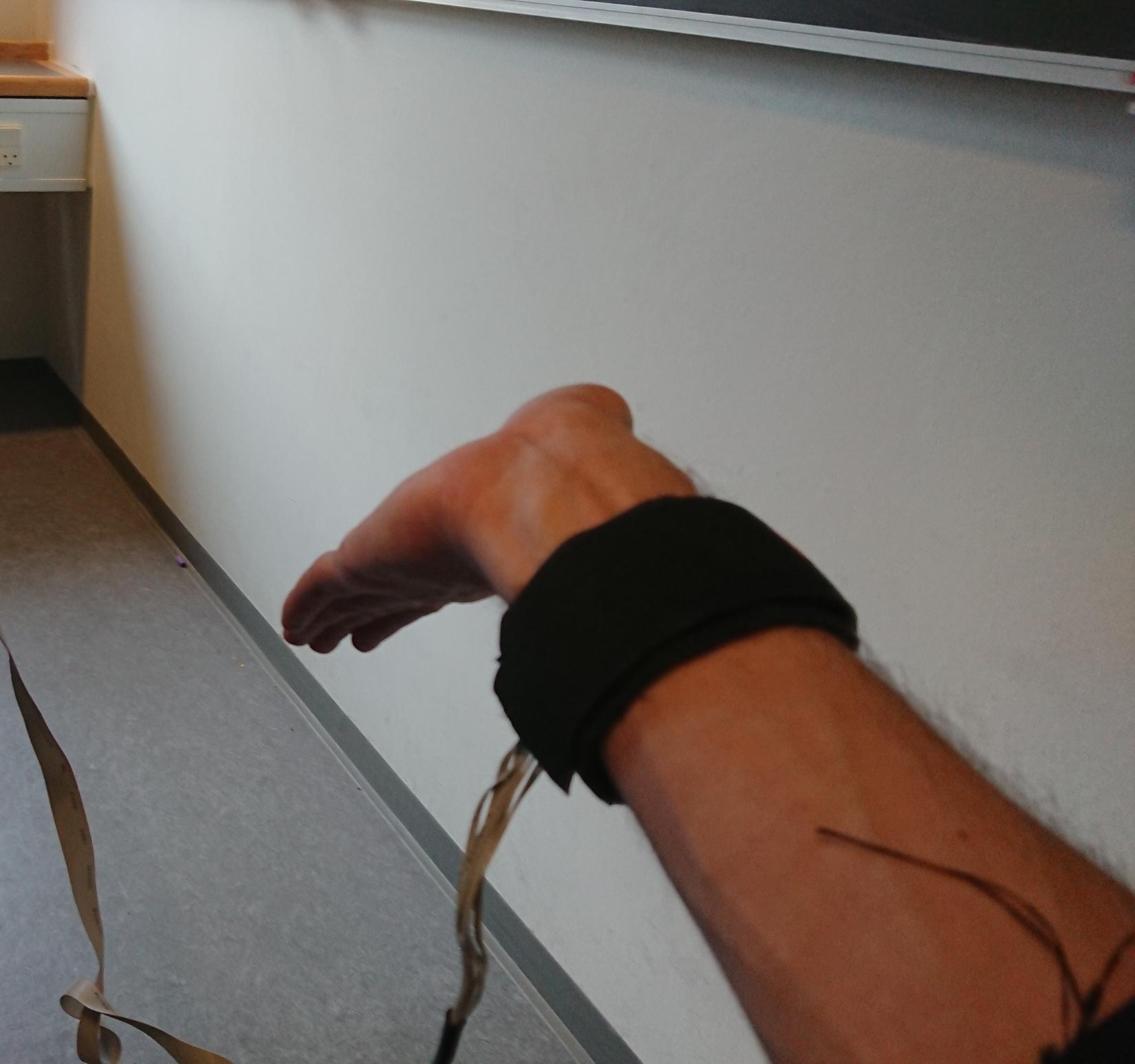}
        \subcaption{\\Supine extension}
        \label{fig:supine_extension}
    \end{minipage}
    \\[0.25cm]
    \begin{minipage}[t]{0.16\linewidth}
        \includegraphics[width=\textwidth]{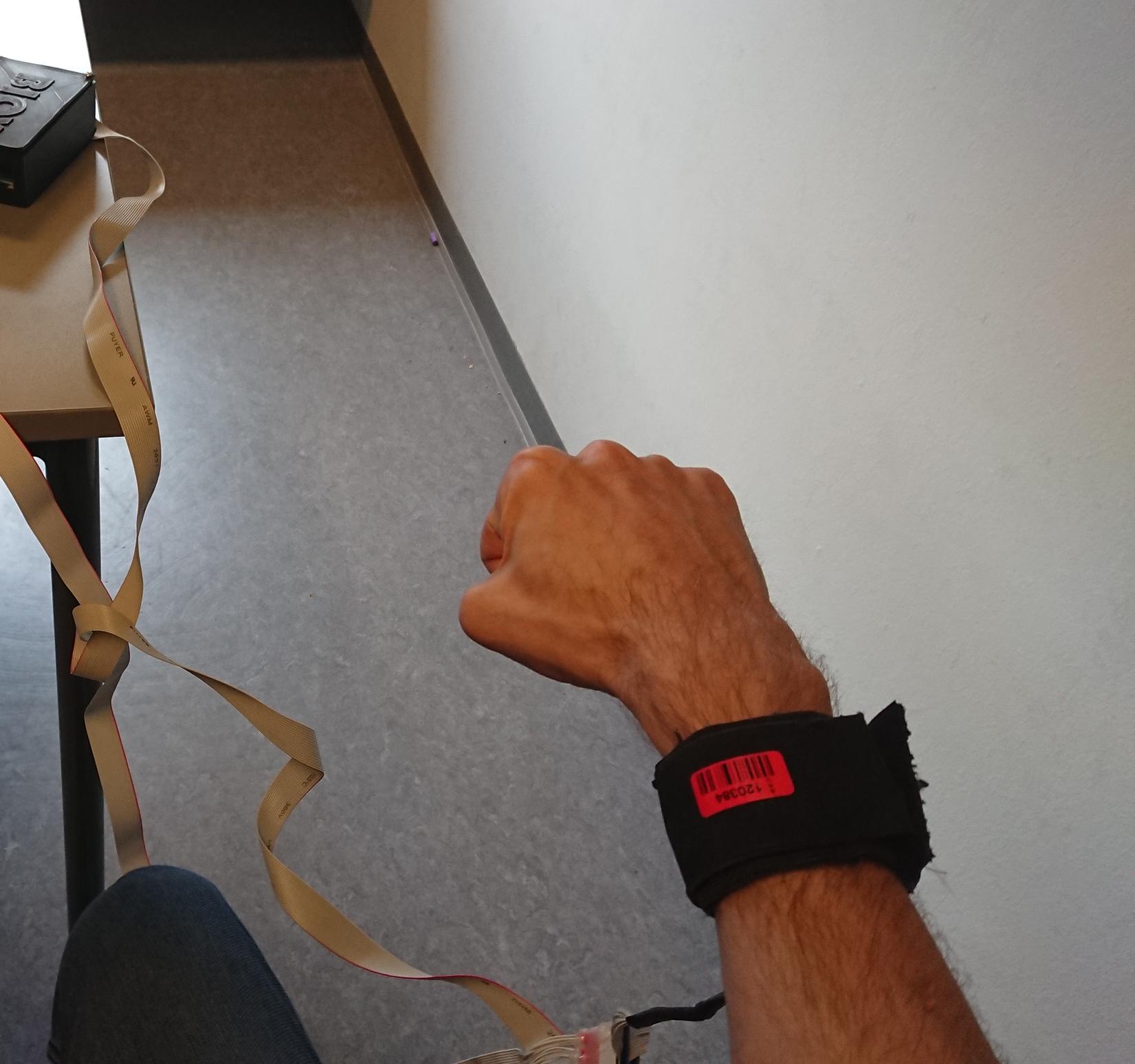}
        \subcaption{\\Prone closed}
        \label{fig:prone_closed}
    \end{minipage}
    \begin{minipage}[t]{0.16\linewidth}
        \includegraphics[width=\textwidth]{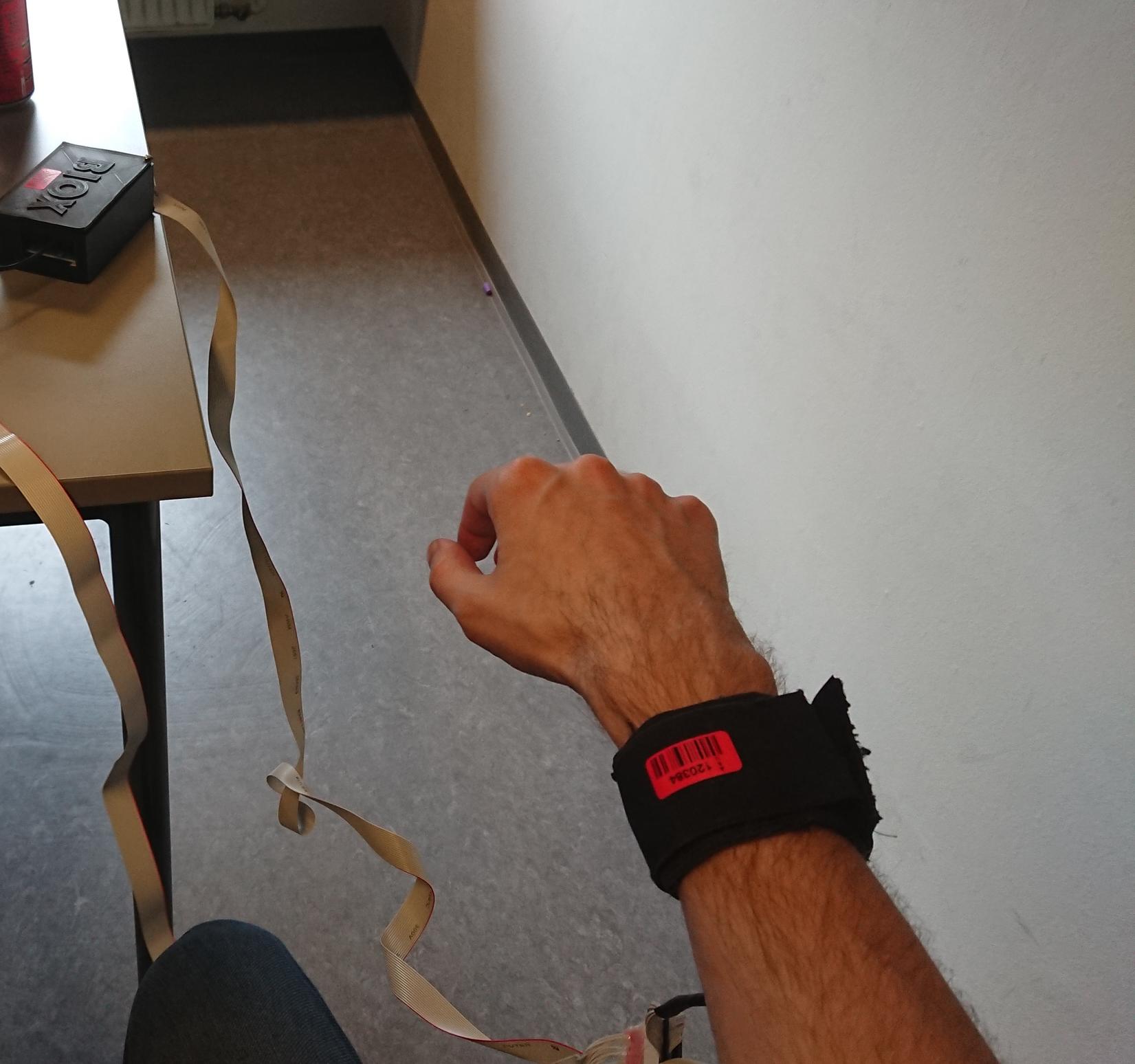}
        \subcaption{\\Prone rest}
        \label{fig:prone_rest}
    \end{minipage}
    \begin{minipage}[t]{0.16\linewidth}
        \includegraphics[width=\textwidth]{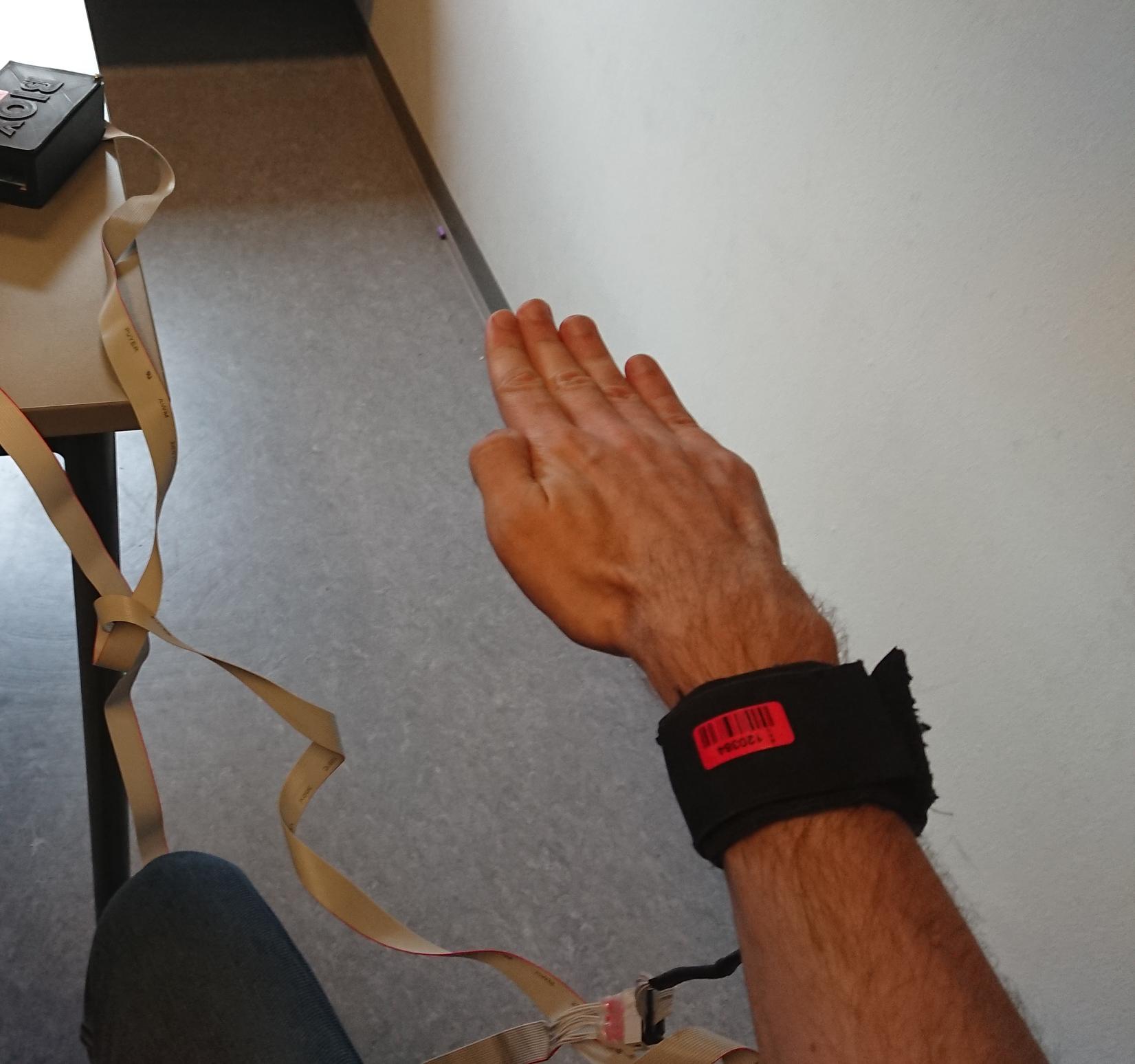}
        \subcaption{\\Prone open}
        \label{fig:prone_straight}
    \end{minipage}
    \begin{minipage}[t]{0.16\linewidth}
        \includegraphics[width=\textwidth]{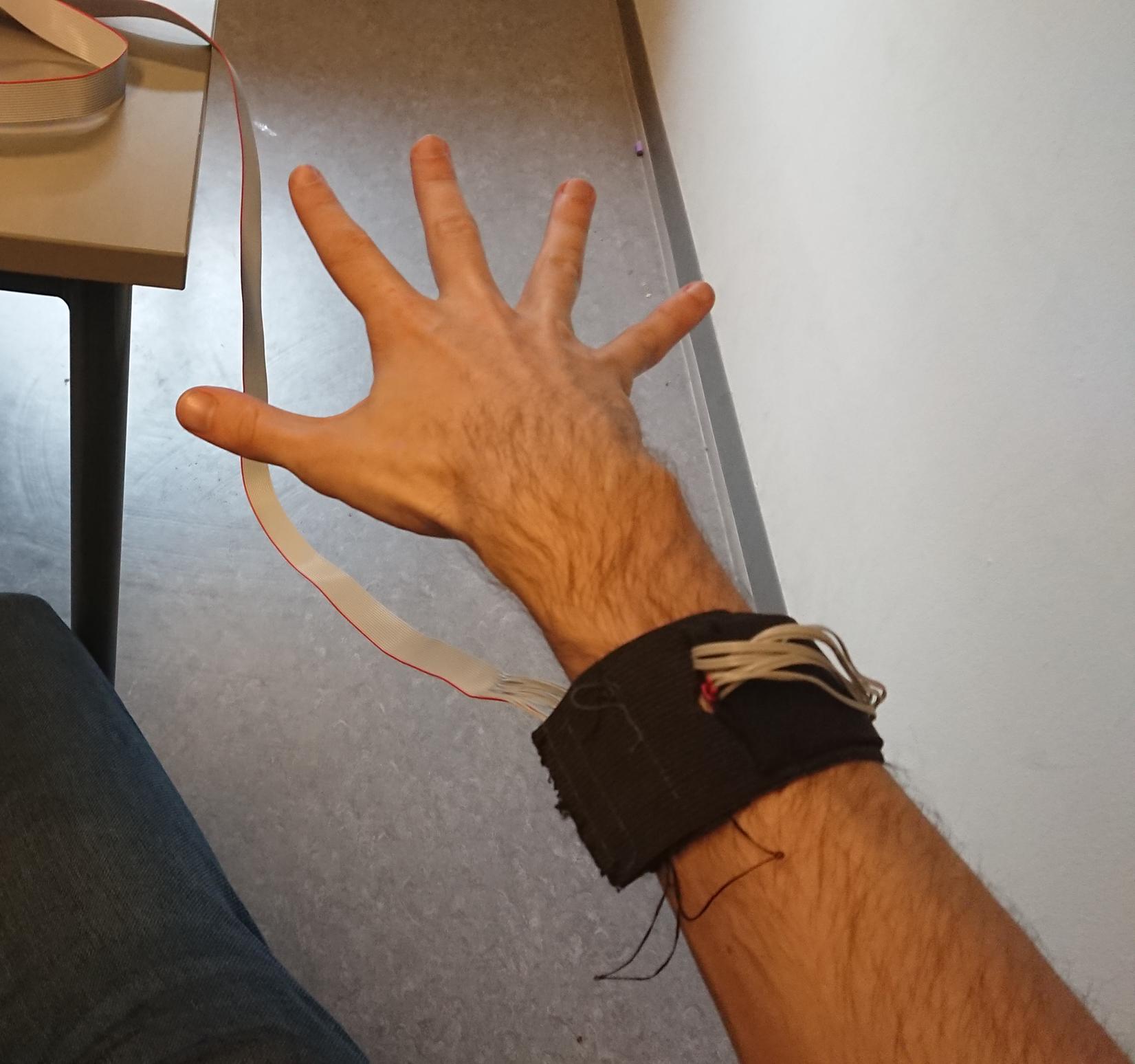}
        \subcaption{\\Prone wide}
        \label{fig:prone_wide}
    \end{minipage}
    \begin{minipage}[t]{0.16\linewidth}
        \includegraphics[width=\textwidth]{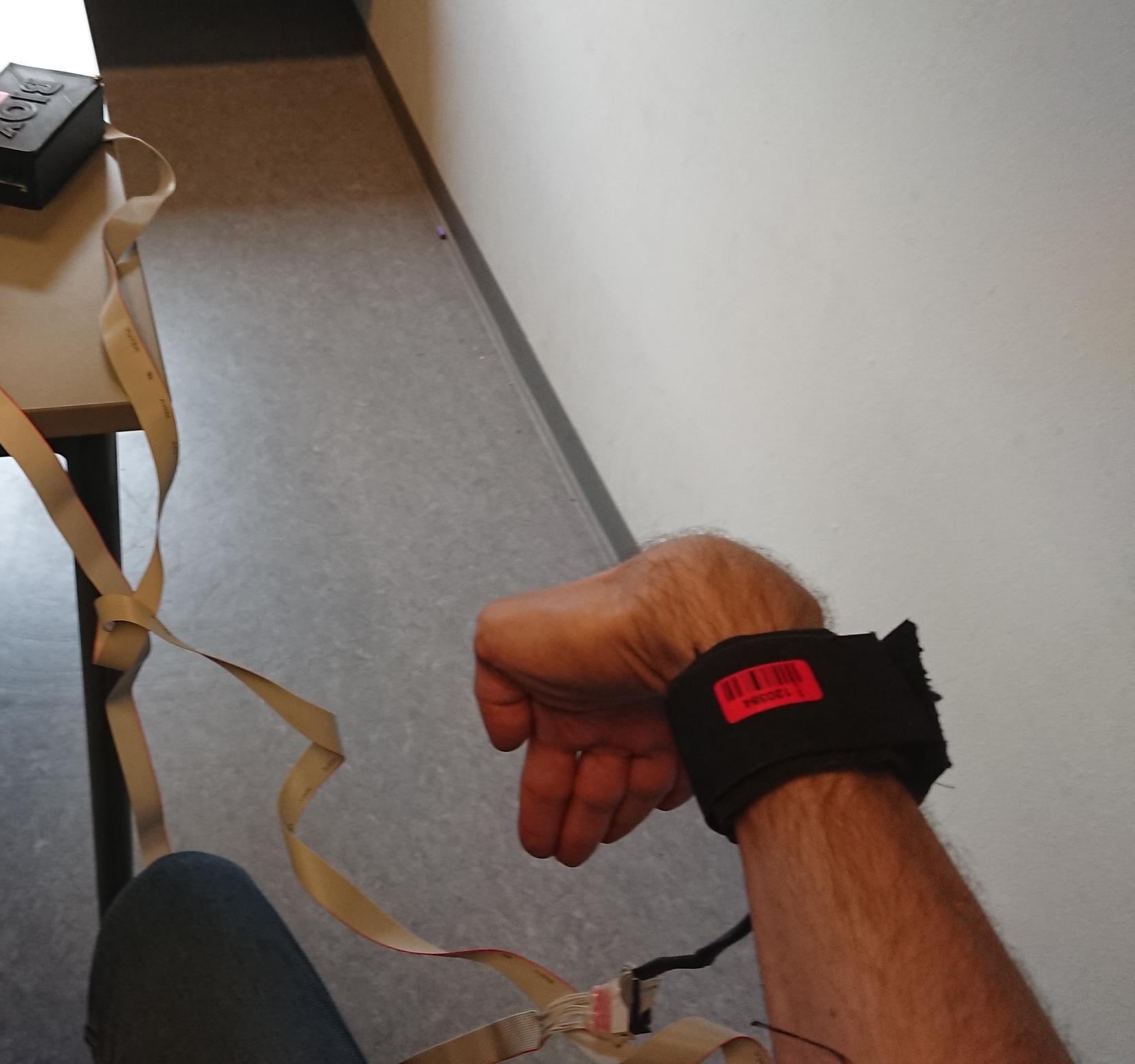}
        \subcaption{\\Prone flexion}
        \label{fig:prone_flexion}
    \end{minipage}
    \begin{minipage}[t]{0.16\linewidth}
        \includegraphics[width=\textwidth]{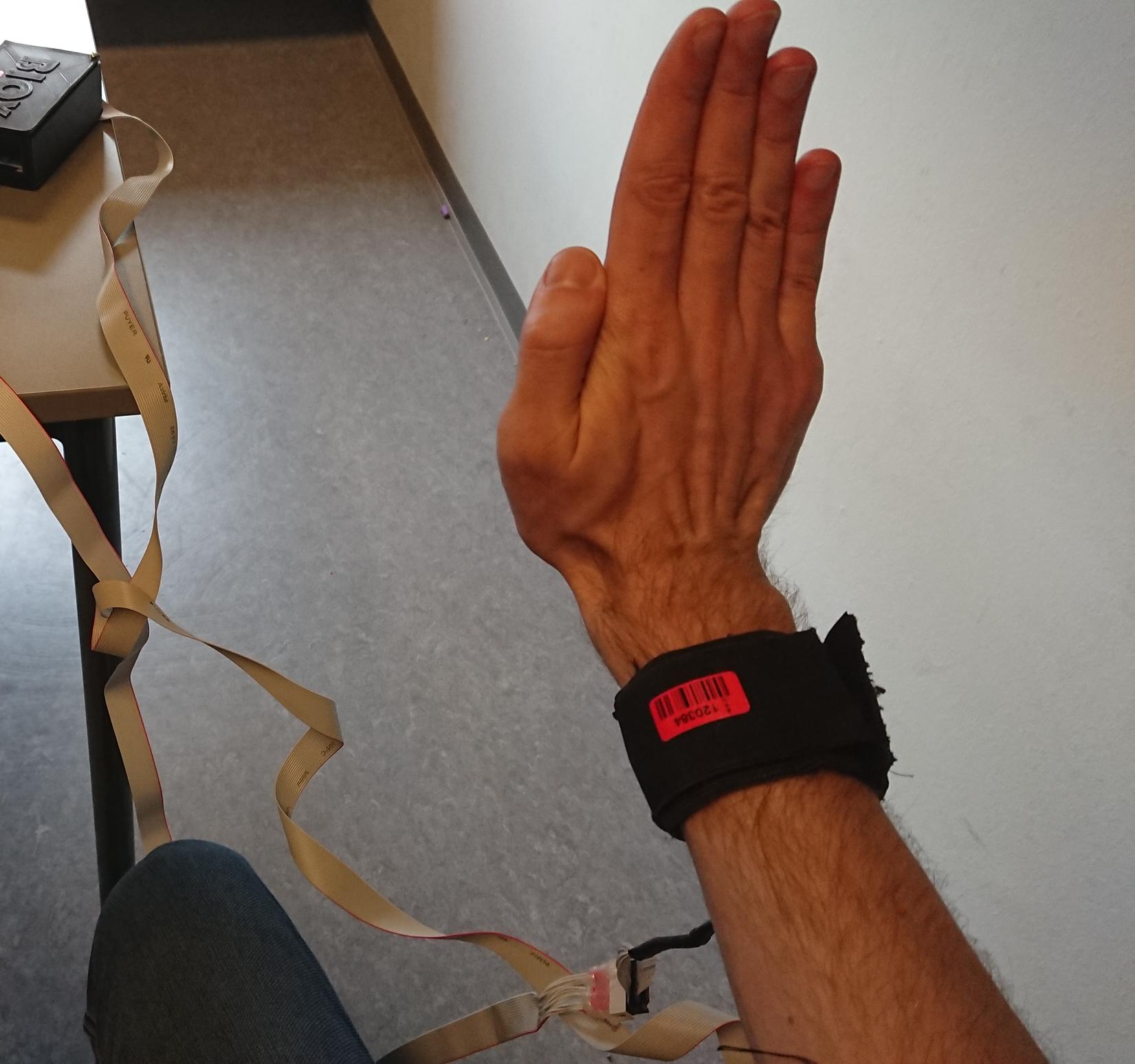}
        \subcaption{\\Prone extension}
        \label{fig:prone_extension}
    \end{minipage}
    \\[0.25cm]
    \caption{Gestures}
    \label{fig:gestures}
\end{figure*}

\subsection{Equipment}
For this study we have used a setup of 2 BIOX Armbands\footnote{\url{https://www.bioxgroup.dk/product/biox-armband/}}, one with 7 sensors for the wrist and a larger one with 8 sensors for the forearm. Both sensors were connected to our laptop, and during data collection we gathered sensor readings from both sensors with a frequency of \varfrequency{}, limited by our laptops processing speed. We sampled at as high frequency as possible to avoid limiting the potential applications, as users can down sample to a lower frequency as appropriate for their application.

\subsection{Consent}
All data was gathered from volunteers. Before any data was collected, subjects were presented with a disclaimer informing them of what data we would collect and how we intended to use it, including informing them of our intent to make the data publicly available: 
\begin{quote}
    We are a research group at Aalborg University attempting to break new ground in the field of gesture recognition, and we need your data to do so. We will collect data such as your age, gender, fitness as well as record your arm / wrist while you do different gestures. That is to say, we will not collect any personally identifiable information (PII) such as your name, address etc. By participating in our data collection, you agree to have your data shared in a public datset. We share the data such that our results may be reproduced and improved upon in the future.
\end{quote}

\subsection{Subject information}
The subjects were presented with a form and asked to provide some contextual information that we expect to have some impact on the gesture identification process. As part of this, we measure the circumference of wrist and arm at the locations we will apply the sensor armbands. For the armband with 7 sensors, this is $\sim$ 5 cm below the wrist, while for the 8 sensor armband it is below the elbow at the maximal bulge of the forearm. The contextual information is described below and the distribution of the data is noted in \autoref{tab:participant_distribution}.

\begin{description}
\item[Age] Research has shown that ageing reduces joint mobility~\cite{joint_mobility_age} which we expect to impact subjects performance of the requested gestures. Additionally age has effects on the mechanical properties of soft tissue (e.g. elasticity, density, thickness)~\cite{physical_aging} which we also expect to affect gesture identification performance.

\item[Gender] Research shows that there are physiological differences~\cite{difference_physiological_gender} including variation in wrist joint mobility~\cite{difference_joint_mobility_gender,difference_wrist_gender} between genders. As such, we expect that the differences will affect gesture identification. 

\item[Handedness] As the joint mobility might be better in the dominant hand.

\item[Weekly Exercises] As what we measure is the mechanical activation of the muscles we expect that physical fitness will impact the readings. We have therefore asked each subject to give the number of days per week when they exercise for at least half an hour.

\item[Injury] If a subject has an injury or condition that affects the mobility of the wrist or forearm this will likely impact the gesture identification. As such we have asked the subjects whether they have an injury that affects wrist or hand mobility.

\item[Wrist and Forearm circumference] We expect that the wrist and forearm circumference will affect gesture identification as it will impact how well the sensor armbands fit the subject.
\end{description}

\begin{table}[htbp]
\centering
\caption{Subject distribution. For numeric properties, $mean (\pm \gls{sd})$.}
\begin{tabular}{|l|l|}
\hline
Gender                & \varmale{} Male, \varfemale{} Female        \\ \hline
Age                   & $\varmeanage{} (\pm \varsdage{})$           \\ \hline
Handedness            & \varrighthanded{} Right, \varlefthanded{} Left \\ \hline
Weekly Exercise              & $\varmeanexercise{} (\pm \varsdexercise{})$ \\ \hline
Injury                & \varinjured{} Yes, \varnoninjured{} No      \\ \hline
Wrist circumference   & $\varmeanwrist{} (\pm \varsdwrist{})$       \\ \hline
Forearm circumference & $\varmeanforearm{} (\pm \varsdforearm{})$   \\ \hline
\end{tabular}%
\label{tab:participant_distribution}
\end{table}

\subsection{Sensor fitting}
We then apply the BIOX armbands at the wrist and forearm. 
The armband is laid on a flat surface  with the cables to the left and the subjects are asked to lay their right forearm on the middle of the armband with the wrist prone, such that the armband match the measured location.
Once the subjects have positioned the arm correctly we close the armbands.

The subjects are then told to sit such that there is $\sim$ 1 m free space in front of them and $\sim$ 0.5 m to their sides and back. They are then asked to position their arm so the upper arm is parallel to the body, the elbow does not touch their side, and their forearm is perpendicular to the upper arm.
They are asked to hold their elbow as still as possible and keep their forearm horizontal and straight in front of them.

When they are positioned correctly they are given a remote controller and instructed to follow the prompts on the screen and click the remote when they have assumed the shown gesture, and then hold the position shown until collection is done and a new prompt is given. 
During the calibration and collection processes we will keep watch to ensure correct execution of the gestures. Should we see any errors we will intervene and ask them to redo the gesture and giving directions for correct execution.

\subsection{Calibration}\label{calibration}
Calibration is performed because the sensors will likely only utilise part of the possible output range. We thus perform calibration in order to better utilise the full output range from the sensors. In order to capture the upper limit of the subjects muscle activation, the subjects were told that they should try to exert their muscles as much as possible when the sensors were being calibrated. A side-effect of calibration is that the resting value\footnote{The resting value is the value the sensors outputs when no pressure is exerted on them.} of the sensors is also increased, depending on the number of calibration steps (i.e.,  how much the sensitivity is increased). 

Since the two sensor armbands are activated differently for each gesture, they are calibrated separately using the two gestures that we found to best activate the sensors of the respective armband. These gestures were \autoref{fig:neutral_closed} and \autoref{fig:prone_extension} for the arm and wrist, respectively. Since the armbands are calibrated separately they will likely require a different number of calibration steps, leading to different resting values. For this reason we have for each subject recorded the number of calibration iterations for each armband as well as the final sensor values on calibration in the belief that this could be used to account for these factors.

\subsection{Data collection}\label{data_collection_process}
The subjects are instructed that they do not need to exert maximum force for the rest, and that they should proceed at their own pace, taking rests as necessary between gestures.

We collect data for a set of 6 hand gestures with 3 different wrist positions, for a total of 18 different combinations of hand and wrist positions which can be seen in \cref{fig:gestures}.  We collect data for each gesture \varnumrepetitions{} times, each time collecting \vargestureduration{} of sensor readings at a frequency of \varfrequency{}.

In addition to the value of each sensor we record a label signifying which gesture was performed, as well as which subject the reading is from. Each reading has a timestamp, though only the difference in time between sensor readings of the same subject is valid, not the time of day due to the implementation of the timer. As we collected data over several repetitions of the gestures we have also included a numeric indicator of which repetition the reading is from. 

The collected data is of the following form:
\begin{align*}
\langle &arm\_sensors, wrist\_sensors, timestamp, \\
&repetition, subjectID, gestureID \rangle
\end{align*}
where $arm\_sensors$ is a 8 dimensional vector and $wrist\_sensors$ is a 7 dimensional vector of sensor readings. When considering a sequence of such data along the time dimension, it gives multi-dimensional time series data~\cite{DBLP:conf/mdm/Kieu0J18,DBLP:journals/vldb/HuYGJ18}. 

\subsection{Risks and Assumptions}
There are a couple of areas of potential risk with our data collection protocol. Firstly, with respect to the contextual information, we relied on the subjects to provide the information which may have inaccuracies, especially on fields such as frequency of exercise where subjects may have embellished their details.

Further, there is some degree of inaccuracy in regards to the wrist and arm measurements. Though we tried to measure as consistently as possible, when dealing with something as inherently soft as an arm it is difficult to manually measure at exactly the same tightness each time.

The same applies to the fitting, which while we strived for consistency likely exhibits some degree of variance, which could lead to difference in the needed calibration and subsequent data collection. 
As described in \autoref{calibration}, we have included the calibration information in order to alleviate this issue.

In regards to the gesture data collection, we chose to let the subjects determine their own pace, signifying when they had assumed the next gesture. While we supervised the data collection, and asked the subjects to redo any gestures where we observed errors, it is possible that some subjects may have pushed the button a bit too early. 

Additionally since we left it up to the subjects to decide when and how long they needed to rest between gestures there may be some variance in their fatigue levels throughout the data collection.
\section{Example use-case}
One use of \gls{fmg} gesture recognition might be to control an exoskeleton, such as for assisting patients during rehabilitation. In such a case, it might be infeasible to collect large amounts of data from the patient. As such it would be helpful if we could use data gathered from other people to supplement the patients data through transfer learning.

Transfer learning uses domains $D$ and tasks $T$, where in $D$ we have a source domain $D_s$, and a target domain $D_t$ for which we want to transfer knowledge to. A domain consists of a feature space $X$ and a marginal probability distribution $P(X)$, where $X= \{x_1,...,x_n\}$ and a task $T = \{Y,f(\cdot)\}$ where $f(\cdot)$ is a predictive function and $Y$ the label space~\cite{transfer_learning_survey}. There are various approaches to transfer learning depending on how knowledge can be transferred effectively from source domain to target domain~\cite{transfer_learning_survey}.

We utilise domain adaptive transfer learning, where we assume that $D_s$ and $D_t$ are different but $T_s$ and $T_t$ are the same~\cite{transfer_learning_survey}, which is appropriate for our problem, as we have different domains (i.e. different subjects) but the gesture recognition task for all domains is to recognise the gestures shown in \autoref{fig:gestures}.

\subsection{Feature Engineering}\label{sec:feature_extracting}
For this use-case example we scale the dataset to take into account how the calibration functions, so as to normalize the data while considering the \textit{resting} values of the sensors. 

Let $D$ be the raw dataset collected for a given subject, consisting of 15 dimensional vectors containing a sensor reading from each of the sensors (7 wrist + 8 arm sensors). Let $d^{i}$ be the set of data readings from sensor $i$, and $d_j$ be the $j$th sample of $D$. The $j$th reading of sensor $i$ is thus $d_{j}^{i}$. For each $d_{j}^{i} \in D$ we subtract the minimum reading for the corresponding sensor, $min(d^{i})$, and divide by the overall maximum reading across every sensor $max(D) = max(\ \{ max(d^{i}) | i \in \{1, ..., 15\} \}\ )$ which is also adjusted by $-min(d^{i})$.

\begin{equation}\label{eq:scaling}
Scale(\:d^{i}_{j})\:=\: \frac{d^{i}_{j}-min(d^{i})}{max(D)-min(d^{i})} \end{equation}


We consider the local minimum of each sensor, as the resting values may be different due to calibration (see \autoref{sec:data_collection}). We are also interested in preserving the relative values between the sensors, we therefore consider the global maximum of the sensors, as the amount of force exerted vary when performing different gestures.
\subsection{Models}\label{sec:model}
We experiment with different deep learning models and evaluate the potential of applying transfer learning to this task.

\begin{figure}
\centering
\begin{minipage}[t]{0.30\linewidth}
\includegraphics[width=\linewidth]{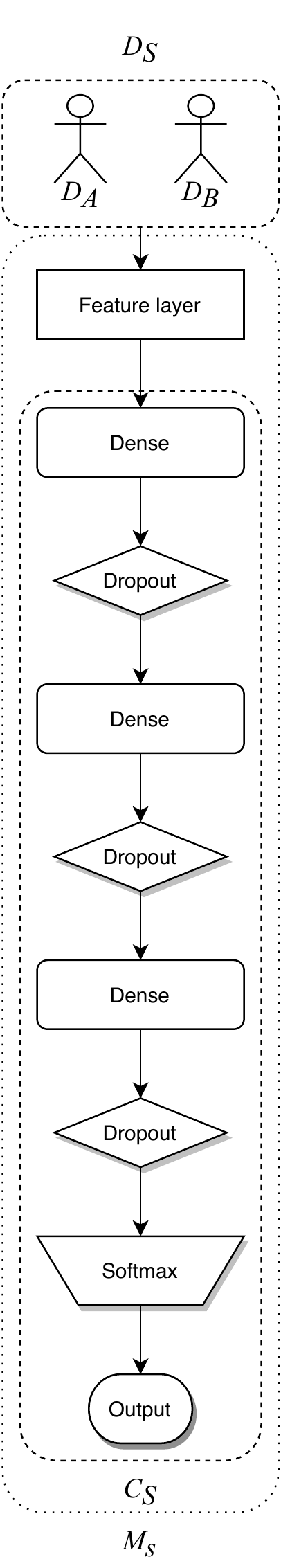}\\
\subcaption{Source model}
\label{fig:cpnna}
\end{minipage}%
\hfill
\begin{minipage}[t]{0.60\linewidth}
\includegraphics[width=\linewidth]{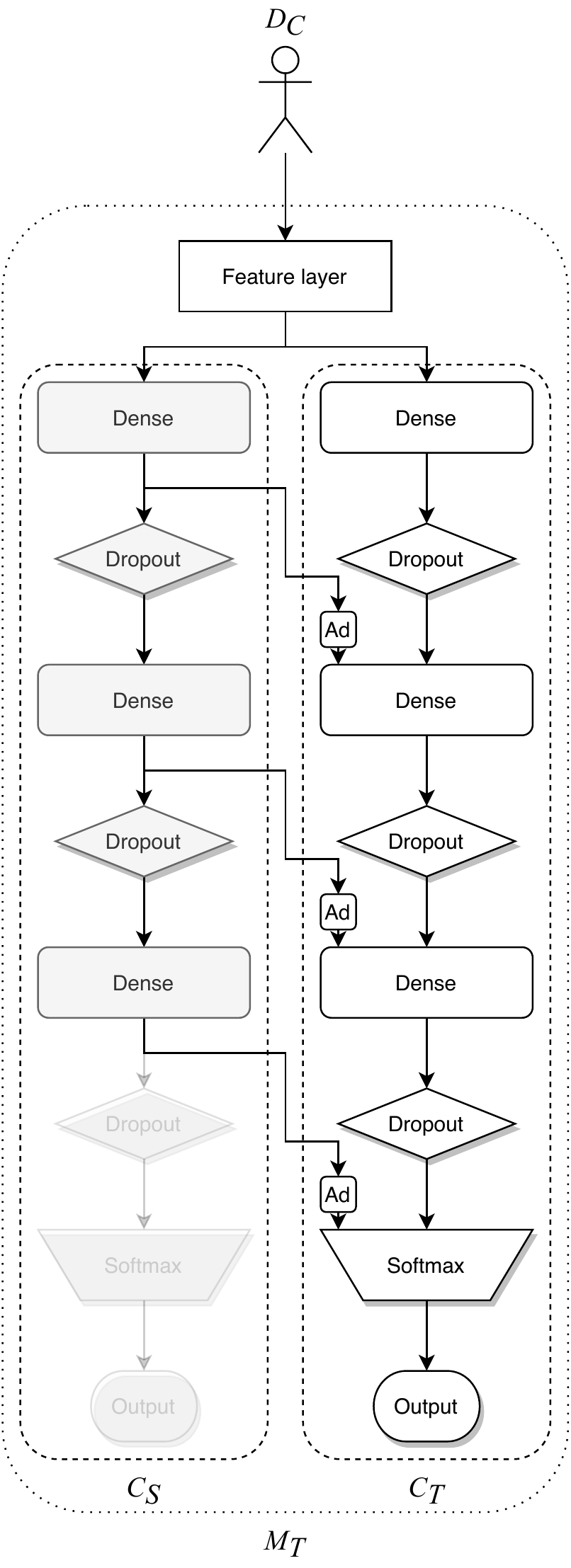}\\
\subcaption{Target model}
\label{fig:cpnnb}
\end{minipage}%
\caption{CPNN Overview. Given 3 subjects with domains $D_A$, $D_B$ and $D_C$, assume  $D_A,\:D_B\in D_S$ belong the source domain and $D_C=D_T$ is the target domain. All source domains in $D_S$ are combined to train a single column $C_S$ in (a). Then that columns weights are frozen and in (b) a new column $C_T$ is created with lateral connections from $C_S$ through adaptors $Ad$. This model is then trained on the target domain $D_C$. As such, regardless of the number of source domains, the size of the model does not change.}
\vspace{-10pt}
\end{figure}

\subsubsection{Baseline}\label{sec:baseline}
Our non-transfer learning baseline simply consists of a series of \gls{fcnn} layers, using the \gls{relu} activation function, followed with a Softmax layer. In order to avoid overfitting we apply dropout between each layer. This architecture also serves as the basis for the columns of our transfer learning approach. As we consider a non knowledge transfer setup, we train a baseline model for each subject.

\subsubsection{Transfer Learning}\label{sec:transfer}
\gls{pnn} were proposed by \cite{progressive} as a way of applying transfer learning to a sequence of tasks while avoiding catastrophic forgetting. 
It works by training on the source domains in sequence, and utilizing these through lateral connections to later models. 
However, as the \gls{pnn} architecture exhibits quadratic growth in the number of parameters when increasing the number of source domains~\cite{progressive}, there is a limit to the number of source domains we can reasonably draw on. Hence, if we want to be able to learn from a large set of source domains we need an alternative to having a column for each. One possible approach which was proposed by \cite{CNN_transfer_learning_sEMG_2019} is to combine the source domain datasets and only train a single column on this combined source domain as seen in \autoref{fig:cpnna}, which like with \gls{pnn} is connected to the target column with lateral connections, as seen in \autoref{fig:cpnnb}. Combining the source domains is possible because, unlike what the original \gls{pnn} was proposed for~\cite{progressive}, we do not have different tasks in addition to the different domains. We can thus combine the source domains and train a single column to learn the general features across the source domains which are helpful for our task. It is thus possible to draw on a large number of source domains without increasing the number of parameters of the model.

\subsection{Training and Evaluation}\label{sec:training_and_evaluation}

As mentioned in \autoref{sec:data_collection}, we have collected data for a series of repetitions, each repetition covering 5 seconds of data for every gesture. Of the 5 total repetitions we have collected for each subject we will use the last repetition for testing and the remaining repetitions for training. We believe this best fits a case where a person may equip the device and collect some initial data to train a model that should then perform predictions on subsequent data. For each of the models, we split the training data into a training and validation set $75/25\%$ and utilise early stopping monitoring the validation loss delta with a patience of $5$ and delta threshold of $0.001$ to determine when to stop training.

\subsubsection{\gls{fcnn}}
For the baseline \gls{fcnn} model, we train a model for each subject using the data collected for the last repetition as test data and the remainder as training data.

\subsubsection{\gls{cpnn}}
During the training of \gls{cpnn} we start by combining the data for all subjects except the one we want to train the current model for. This combined dataset includes all 5 repetitions for these subjects. With this combined dataset we pretrain a model as described in \cref{sec:model}. We then take the data for the target subject and separate the last repetition for testing as mentioned above, before training on the remaining repetitions.
\subsection{Results}\label{results}
Based on three model configurations, we fit \gls{fcnn} and \gls{cpnn} models. Based on the best and worst subject models, derived from the models with the highest mean accuracy, we construct a confusion matrix such that we can observe the classifications differences.
\begin{table}[htbp]
\caption{Evaluation Results. Mean accuracy $\mu$ and standard deviation $\sigma$. Optimal number of neurons per layer and dropout rate were identified by hyperparameter optimisation.}
\resizebox{\linewidth}{!}{%
\newcounter{rowcount}
\setcounter{rowcount}{0}
\newcommand\rownumber{\stepcounter{rowcount}\arabic{rowcount}}
\centering
\begin{tabular}{ccccccccc}
\toprule
\#         & Dense 1 & Dense 2 & Dense 3 & Dropout & \multicolumn{2}{c}{FCNN}    & \multicolumn{2}{c}{CPNN} \\
           &         &         &         &         & $\mu$            & $\sigma$ & $\mu$            & $\sigma$  \\ \midrule
\rownumber & 256     & -       & -       & 0.4     & 77.59\%          & 13.20\%  & \textbf{78.12}\% & 12.83\%    \\
\rownumber & 256     & 32      & -       & 0.1     & 77.81\%          & 13.06\%  & \textbf{78.28}\% & 12.88\%    \\
\rownumber & 256     & 64      & 32      & 0.3     & \textbf{78.14}\% & 12.63\%  & 77.11\%          & 14.01\%   \\ \bottomrule
\end{tabular}%
}
\label{tab:all_results}
\end{table}

\subsubsection{\gls{fcnn}}
\begin{figure*}[htbp]
    \centering
    \begin{minipage}[t]{0.49\linewidth}
        \includegraphics[width=\linewidth]{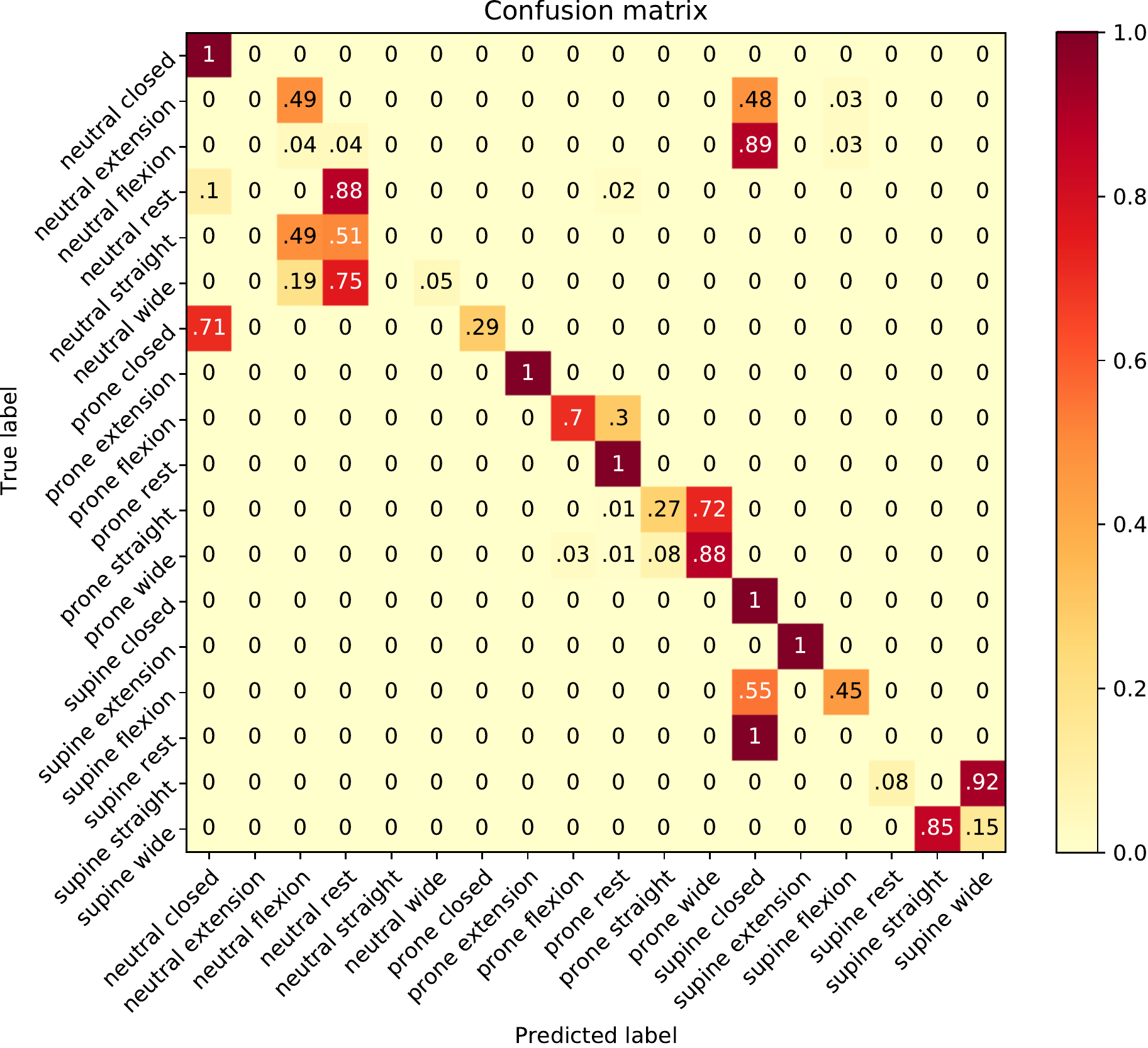}
        \subcaption{Worst performing subject, subject id 15. Accuracy 48.39\%.}
        \label{fig:fcnn_worst}
    \end{minipage}
    \hfill
    \begin{minipage}[t]{0.49\linewidth}
        \includegraphics[width=\textwidth]{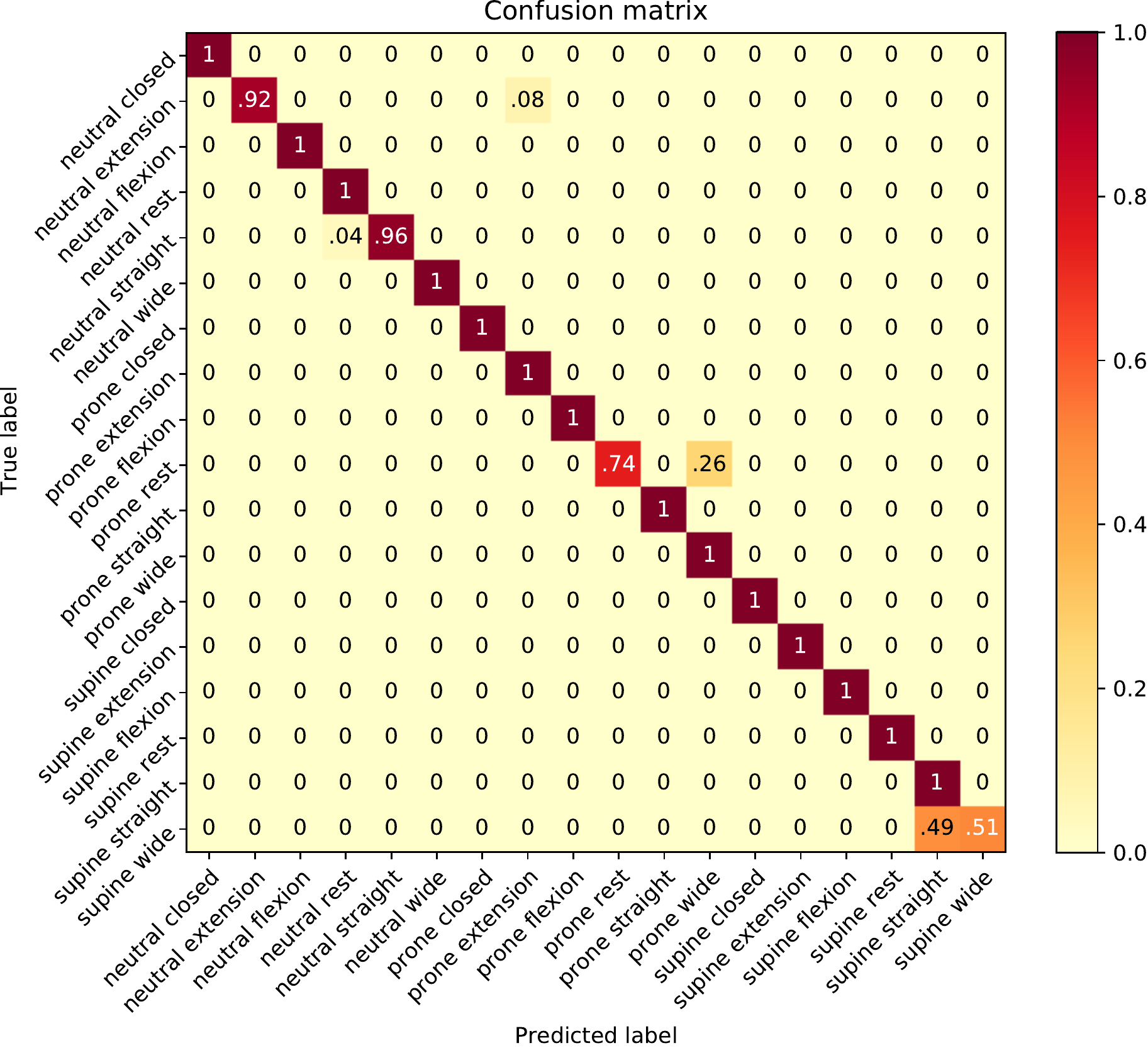}
        \subcaption{Best performing subject, subject id 18. Accuracy 95.21\%}
        \label{fig:fcnn_best}
    \end{minipage}
    \caption{Confusion matrix for best performing baseline \gls{fcnn} based on our results in \autoref{tab:all_results}.}\label{fig:cm_fcnn}
\end{figure*}

Looking at the test results in \autoref{tab:all_results}, we observe that deeper models outperform the shallower ones. 

Looking at the confusion matrix for our best and worst performing \gls{fcnn} subject models in \autoref{fig:cm_fcnn}, we observe some interesting misclassifications. Looking at the confusion matrix for our worst performing \gls{fcnn} model in \autoref{fig:fcnn_worst}, we observe that the gesture \textit{supine closed} is predicted correctly, however the model cannot distinguish between \textit{supine closed} and \textit{supine rest}, both of which it classifies as \textit{supine closed}. Likewise in the confusion matrix for our best performing \gls{fcnn} model in \autoref{fig:fcnn_best}, we observe that the model cannot distinguish between \textit{supine straight} and \textit{supine wide} as well as \textit{prone rest} and \textit{prone wide}.

\subsubsection{\gls{cpnn}}
We base our \gls{cpnn} models on the best performing baseline \gls{fcnn} architectures, from our \gls{hpo}. \gls{cpnn} accuracy for the first two model configurations, shows a higher mean accuracy than \gls{fcnn}, however the mean accuracy for \gls{cpnn} when using three layers is worse than that of \gls{fcnn}.

\begin{figure*}[htbp]
    \centering
    \begin{minipage}[t]{0.49\linewidth}
        \includegraphics[width=\linewidth]{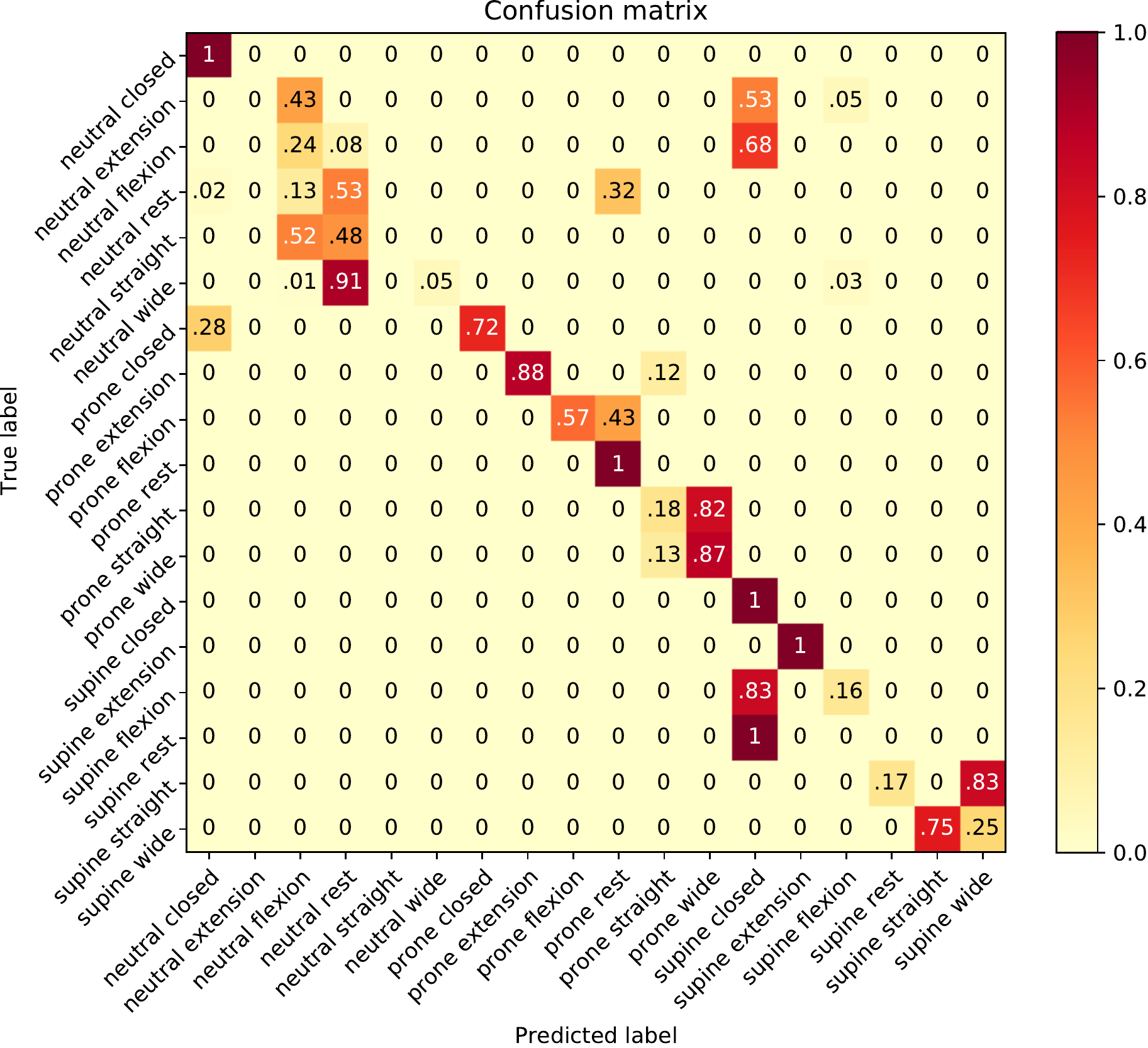}
        \subcaption{Worst performing subject, subject id 15. Accuracy 47.00\%.}
        \label{fig:cpnn_worst}
    \end{minipage}
    \hfill
    \begin{minipage}[t]{0.49\linewidth}
        \includegraphics[width=\linewidth]{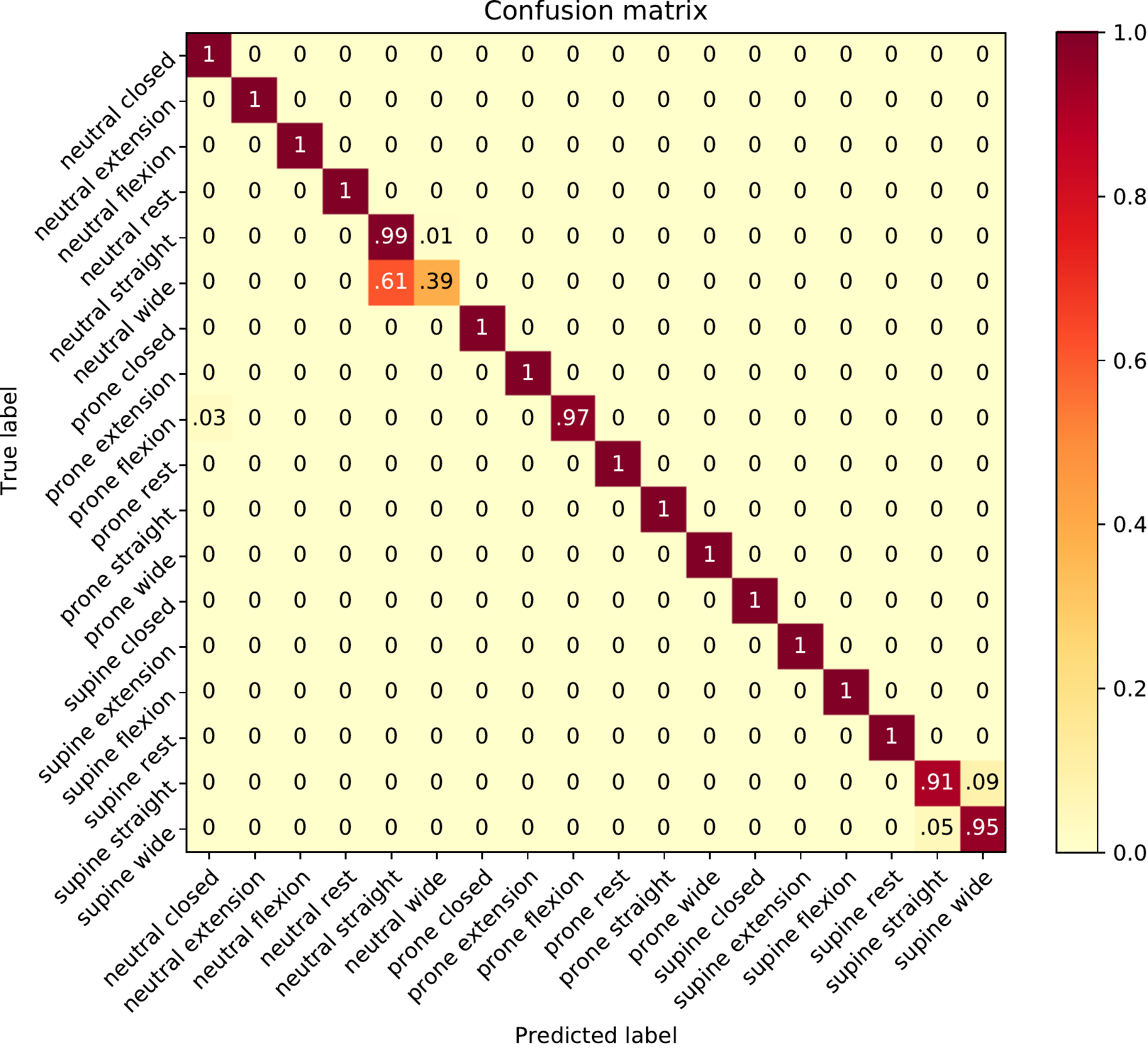}
        \subcaption{Best performing subject, subject id 16. Accuracy 95.61\%}
        \label{fig:cpnn_best}
    \end{minipage}
    \caption{Confusion matrix for best performing baseline \gls{cpnn} based on our results in \autoref{tab:all_results}.}\label{fig:cm_cpnn}
\end{figure*}

Looking at the confusion matrix for our best and worst performing \gls{cpnn} subject models in \autoref{fig:cm_cpnn} and compare with \autoref{fig:cm_fcnn}, we observe some interesting differences. The worst performing subject model correctly classifies \textit{prone closed}, but has a harder time classifying gestures such as \textit{prone flexion} and \textit{neutral rest}. The best performing subject model has a hard time classifying \textit{neutral wide} as it is often misclassified with \textit{neutral straight}. 


\section{Discussion}\label{sec:discussion}

In this section, we will discuss our use of and the potential use of our benchmark dataset, as well as future plans for the dataset.

\subsection{Transfer learning}
Our transfer learning model, \gls{cpnn}, performed slightly better than our baseline \gls{fcnn} in two out of three cases. This suggests that there is potential for knowledge transfer but that one should be careful when applying transfer learning lest one end up with negative transfer. We thus hope that this dataset may also serve as a resource for further transfer learning and multi-task learning research on this domain.

\subsection{Data collection}
While we provide a dataset collected from 20 subjects including some contextual information, those 20 subjects were drawn from a a relatively narrow demographic. We had hoped to collect data more extensively from other segments of the population to achieve a more diverse dataset, especially in regards to the subjects age and gender. However, due to the circumstances surrounding the Covid-19 pandemic and the resulting quarantine\footnote{\url{https://politi.dk/en/coronavirus-in-denmark}}, we had to halt our data collection. 20 subjects is still decent compared to the number of subjects for which data was collected in the related work, mentioned in \autoref{sec:related}, and as such we feel it is still a valuable contribution.

In the future, we would like to extend the dataset to make it more diverse. Furthermore we would like to collect more long term data, such as collecting data from an impaired subject throughout their rehabilitation period.

\section{Conclusion}
We have collected a \gls{fmg} benchmark dataset for hand gesture recognition using a commercially available sensor setup. We have collected benchmark data for \varnumsubjects{} subjects, including contextual information about the subject, for a total of \varnumgestures{} unique gestures. 

The data is collected at a very high frequency up to 1000 Hz, which makes it possible for a wide variety of applications, as users can down sample to a lower frequency as appropriate for their applications. This also provides opportunities for time series analytics, such as prediction~\cite{DBLP:conf/cikm/CirsteaMMG018} and outlier detection~\cite{DBLP:conf/ijcai/KieuYGJ19}. 

We have used this dataset to show that transfer learning has the potential to increase recognition accuracy by incorporating knowledge learned from other subjects. However, negative transfer may happen. Both the dataset and the source code of the use-case have been made publicly available on GitHub~\cite{data}. We believe that this dataset will facilitate research both on \gls{fmg} based hand gesture recognition and on transfer learning.


\bibliographystyle{ACM-Reference-Format}
\bibliography{main}

\appendix

\end{document}